%% file: main.tex
\documentclass{article}

\usepackage{microtype}
\usepackage{graphicx}
\usepackage{subfigure}
\usepackage{booktabs} 

\usepackage{hyperref}

\usepackage[accepted]{icml2025}

\usepackage{amsmath}
\usepackage{amssymb}
\usepackage{mathtools}
\usepackage{amsthm}
\usepackage{subcaption}
\usepackage{graphicx}    
\usepackage{caption}     
\usepackage[capitalize,noabbrev,nameinlink]{cleveref}

\usepackage[normalem]{ulem} 

\theoremstyle{plain}

\theoremstyle{definition}

\theoremstyle{remark}

\newcommand{\Autoref}[1]{%
\begingroup%
\renewcommand{\equationautorefname}{Equation}%
\renewcommand{\figureautorefname}{Figure}%
\autoref{#1}%
\endgroup%
}

\usepackage[textsize=tiny]{todonotes}

\usepackage{verbatim}

\icmltitlerunning{Clustering via Self-Supervised Diffusion}

\begin{document}

\twocolumn[
\icmltitle{Clustering via Self-Supervised Diffusion}



\icmlsetsymbol{equal}{*}

\begin{icmlauthorlist}
\icmlauthor{Roy Uziel}{yyy,aaa}
\icmlauthor{Irit Chelly}{yyy,aaa}
\icmlauthor{Oren Freifeld}{yyy,aaa,zzz}
\icmlauthor{Ari Pakman}{xxx,zzz,aaa}

\end{icmlauthorlist}

\icmlaffiliation{yyy}{Department of Computer Science, Ben-Gurion University of the Negev, Beer Sheva, Israel}
\icmlaffiliation{xxx}{Department of Industrial Engineering and Management, Ben-Gurion University of the Negev, Beer Sheva, Israel}
\icmlaffiliation{zzz}{The School of Brain Sciences and Cognition, Ben-Gurion University of the Negev, Beer Sheva, Israel}
\icmlaffiliation{aaa}{Data Science Research Center, Ben-Gurion University of the Negev, Beer Sheva, Israel}

\icmlcorrespondingauthor{Roy Uziel}{uzielr@post.bgu.ac.il}


\icmlkeywords{Machine Learning, ICML}

\vskip 0.3in
]


\printAffiliationsAndNotice{} 

{}  

\input{sections/abstract}

\input{sections/introduction}

\input{sections/background}

\input{sections/model}

\input{sections/experiments}

\newpage 

\section*{Acknowledgments}
This work was supported in part by the Lynn and William Frankel Center at BGU CS, by Israel Science Foundation Personal Grant \#360/21, and by the Israeli Council for Higher Education (CHE) via the Data Science Research Center at BGU. 
A.P. was supported by the Israel Science Foundation (grant No. 1138/23).
I.C. was also funded in part by the Kreitman School of Advanced Graduate Studies, by BGU’s Hi-Tech Scholarship,
and by the Israel’s Ministry of Technology and Science Aloni Scholarship.

\section*{Impact Statement}
This paper presents work whose goal is to advance the field of Machine Learning. There are many potential societal consequences of our work, none which we feel must be specifically highlighted here.

\bibliography{main}
\bibliographystyle{icml2025}

\newpage
\appendix
\input{images/rescaling_factor_x}
\input{sections/supplementary}
\end{document}

%% file: sections/abstract.tex
\begin{abstract}

Diffusion models, widely recognized for their success in generative tasks, have not yet been  applied to clustering. We introduce Clustering via Diffusion (CLUDI), a self-supervised framework that combines the generative power of diffusion models with pre-trained Vision Transformer features to achieve robust and accurate clustering. CLUDI is trained via a teacher-student paradigm: the teacher uses stochastic diffusion-based sampling to produce diverse cluster assignments, which the student refines into stable predictions. This stochasticity acts as a novel data augmentation strategy, enabling CLUDI to uncover intricate structures in high-dimensional data. Extensive evaluations on challenging datasets demonstrate that CLUDI achieves state-of-the-art performance in unsupervised classification, setting new benchmarks in clustering robustness and adaptability to complex data distributions. Our code is available at \url{https://github.com/BGU-CS-VIL/CLUDI}.
\end{abstract}

%% file: sections/introduction.tex
\section{Introduction}

Clustering is a fundamental task in unsupervised learning, essential for uncovering meaningful groupings within data. These groupings play a vital role in diverse downstream applications, such as image segmentation~\cite{mittal2022comprehensive,friebel2022guided}, anomaly detection~\cite{song2021deep}, and bioinformatics~\cite{karim2021deep}. Despite significant advancements, traditional methods face significant challenges, particularly in datasets with intricate structures and varying intra-class similarity, where such approaches often struggle to capture underlying patterns~\cite{ben2018clustering}. 
\input{images/intro_fig_2}%
To address these limitations, deep learning-based clustering approaches have gained substantial attention for their ability to tackle complex and diverse data landscapes~\cite{zhou:2022:survey1,ren2024deep,wei2024overview}.

However, deep learning-based clustering methods still face persistent challenges that limit their practical utility. A primary issue is \textit{model collapse}~\cite{amrani2022self}, where learned representations degenerate into trivial solutions. Another obstacle arises from underutilizing pre-trained features, resulting in suboptimal performance. 
Notably, \citet{adaloglou2023exploring} demonstrate that using pre-trained Vision Transformers~\cite{caron2021emerging} can surpass methods that attempt to learn both representations and cluster assignments simultaneously, thereby highlighting the effectiveness of high-quality feature initialization.

{Self-supervised representations have consistently outperformed supervised ones in transfer tasks~\cite{ericsson2021well}, highlighting the benefits of leveraging powerful pre-trained models. Meanwhile, many clustering frameworks rely on  complex augmentation pipelines, such as neighbor mining~\cite{van2020scan,adaloglou2023exploring}, or employ {multiple clustering heads} trained in parallel, selecting the best-performing one at evaluation time~\cite{adaloglou2023exploring,amrani2022self}. Although these strategies can improve accuracy, they introduce additional complexity and computational overhead during training.}

To address these limitations, we introduce \textbf{Clustering via Diffusion (CLUDI)}, a self-supervised framework that leverages the generative strengths of diffusion models to produce robust cluster probabilities. As 
\autoref{fig:intro_fig_2} illustrates, CLUDI takes as input a pre-trained Vision Transformer feature vector $\mathbf{x} \in \mathbb{R}^n$ and refines an initial random assignment embedding vector through an iterative diffusion process. Conditioned on $\mathbf{x}$, this process evolves over multiple steps and culminates in a final \textbf{assignment embedding} $\mathbf{z}_0 \in \mathbb{R}^d$, which encodes the likelihood of each data point belonging to each of the $K$ clusters.

During inference, CLUDI generates multiple stochastic samples of $\mathbf{z}_0$ and aggregates the corresponding predictions. By averaging across diverse representations, this strategy mitigates uncertainty, uncovers subtle structures in high-dimensional feature spaces, and yields more stable and accurate cluster assignments, even in complex and diverse data scenarios.

{Our model is trained using a self-supervised Siamese architecture comprising teacher and student branches. The teacher, implemented as a diffusion model, generates assignment embeddings $\mathbf{z}_0 \in \mathbb{R}^d$ and cluster probabilities $p(k|\mathbf{x})$ for $k \in
\{1,2,\ldots,K\}$, through stochastic sampling. These outputs, which capture diverse and complementary views of the data, serve as targets for the student.

The student adapts its predictions to match the teacher’s outputs, enabling it to uncover meaningful and distinct clusters in the data. The training process optimizes two complementary objectives: an \textit{asymmetric non-contrastive loss}~\cite{grill2020bootstrap, chen2021exploring, caron2021emerging} for the assignment embedding $\mathbf{z}_0$ and a \textit{non-collapsing cross-entropy loss} for the cluster probabilities $p(k|\mathbf{x})$. To
facilitate effective clustering and prevent trivial solutions, the cross-entropy loss incorporates a uniform prior over the data mini-batch~\cite{amrani2022self}, encouraging well-separated and diverse cluster assignments.
}

\textbf{Why use diffusion for clustering?}

Despite their success in generative tasks, diffusion models have not been explored for clustering until now. We argue, however, that their ability to model and sample from complex, high-dimensional distributions makes them particularly suited for clustering high-dimensional data, such as images. By iteratively refining noisy representations, diffusion models can uncover underlying structure and variability in the data, providing a natural mechanism for capturing meaningful cluster assignments. Their stochastic nature further enables robust and diverse clustering predictions.

\textbf{Our key contributions are as follows:}
{\begin{itemize}
        \item We introduce \textbf{Clustering via Diffusion (CLUDI)}, a novel framework that is the first to leverage diffusion models for the clustering task.
    \item We propose a \textbf{self-supervised training paradigm} based on a teacher-student architecture, where a diffusion model generates stochastic and informative cluster assignments to guide the student in learning meaningful cluster probabilities.
    \item We demonstrate the effectiveness of \textbf{CLUDI} through extensive evaluations on benchmark datasets, achieving state-of-the-art accuracy and robustness in unsupervised classification tasks.
\end{itemize}}

%% file: images/intro_fig_2.tex
\begin{figure}[!t]
    \centering
\includegraphics[width=0.95\linewidth]{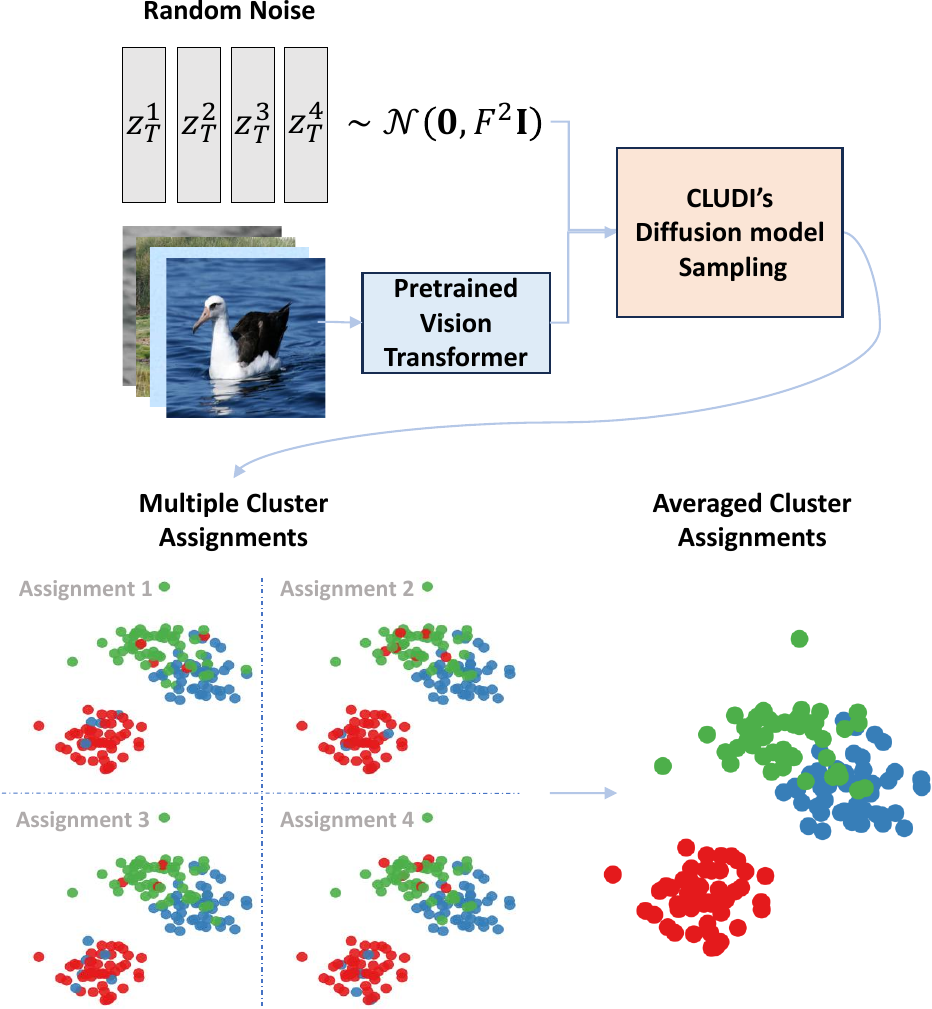}
   \caption{\textbf{Overview of the CLUDI Framework at Inference.}
    Images pass through a pre-trained Vision Transformer to obtain their feature representations.
    Multiple random vectors are sampled from a Gaussian distribution.
    A diffusion model, conditioned on the features,     
    refines the random vectors into class assignment embeddings.
    Each refined embedding corresponds to a candidate cluster probability vector.
     By averaging multiple such assignments, the framework produces robust and accurate clustering predictions.}
\label{fig:intro_fig_2}
\end{figure}

%% file: sections/background.tex
\section{Related Work} 
\label{sec:related}

Deep learning-based models that learn to cluster are usually referred to as performing \textit{deep clustering} or 
\textit{unsupervised classification}. 
Cluster categories are learned during the training phase and remain fixed during inference. These methods have received much interest in the machine learning community.  
Comprehensive reviews on this vast field can be found in~\cite{zhou:2022:survey1,ren2024deep,wei2024overview}.

\paragraph{End-to-end training.} 
Most previous works learn simultaneously 
feature representations and cluster categories. 
An early work in this area is Deep Embedded Clustering (DEC)~\cite{xie2016unsupervised}, which jointly optimizes feature learning through an autoencoder and assigns clusters using a Kullback-Leibler (KL) divergence-based loss. Although DEC can be effective in learning clusters, its performance is sensitive to initialization and prone to model collapse. 

DeepCluster \cite{caron2018deep} takes a different approach by alternating between pseudo-label generation and feature representation refinement. The method iteratively updates cluster centroids and uses the centroids to assign pseudo-labels, which are then used to optimize the feature space. 

However, the computational overhead associated with generating and updating pseudo-labels can be significant, especially for large-scale datasets.

Invariant Information Clustering (IIC) \cite{ji2019invariant} offers a complementary method by maximizing mutual information between different augmentations of the same data. 

Among probabilistic approaches, 
Variational Deep Embedding (VaDE) \cite{jiang2016variational} integrates variational autoencoders with Gaussian mixture models  to learn probabilistic cluster assignments. 

Recent works leverage popular self-supervised approaches. Self-Classifier~\cite{amrani2022self} uses a Siamese network to simultaneously learn representation and cluster labels. To avoid degenerate solutions, it uses a variant of the cross-entropy loss which we adopt in our model. 

\paragraph{Training on pre-trained features.} 
The idea of decoupling feature learning from  cluster learning has been advocated by SCAN~\cite{van2020scan} 
and  TwoStageUC~\cite{han2020mitigating}. 
These models, however, lack efficiency, as they learn features from scratch for every dataset.  

The use of pre-trained features, particularly Vision Transformers (ViTs) \cite{dosovitskiy2020image}, was proposed by TSP~\cite{zhou2022deep} and TEMI~\cite{ren2024deep}. 
These approaches allow the model to focus on refining cluster assignments without relearning low-level features, significantly reducing the computational cost for large-scale datasets.
Our CLUDI approach relies on a similar feature extraction backbone, based on the DINO model~\cite{caron2021emerging}, but proposes a more refined clustering model and training setup that leads to superior results.

DeepDPM~\cite{Ronen:CVPR:2022:DeepDPM}, which can be used in either end-to-end fashion or with pretrained features, is a deep clustering method 
that infers $K$ and is inspired
by a Dirichlet process mixture sampler~\cite{chang2013parallel,dinari2019distributed}. However,
unlike CLUDI, it makes
a stringent assumption
about the distribution 
of the features
within each cluster.

\paragraph{Amortized clustering.}
In another approach, called \textit{amortized clustering}, the model does not learn fixed cluster categories. Instead, it learns  to organize full datasets into clusters discovered at test time.
This approach allows for real-time adaptation of clusters as new data is introduced. It is thus a form of meta-learning~\cite{hospedales2021meta}, 
and research on this task is just beginning to unfold~\cite{pakman2020neural,jurewicz2023catalog,wang2024amortized,chellyconsistent}. 

\section{Background}
\label{sec:background}
\paragraph{Diffusion Models.}

Denoising diffusion probabilistic models (DDPMs) are a recent pivotal shift in the landscape of generative 
modeling~\cite{sohl2015deep,ho2020denoising}, with 
considerable success across a range of applications, including image synthesis, audio generation, and molecular design~\cite{song2020denoising, dhariwal2021diffusion, nichol2021improved, rombach2022high}.

In DDPMs, an initial data sample denoted 
by~\( \mathbf{z}_0 \in \mathbb{R}^d  \),  is transformed 
into pure Gaussian noise, 
\( \mathbf{z}_T \sim \mathcal{N}(\mathbf{0}, \, \mathbf{I}_d), 
\mathbf{z}_T \in \mathbb{R}^d \), through a sequence of incremental additions of Gaussian noise. This  {\it forward process} is  Markovian and defined by:
\begin{equation}
    q(\mathbf{z}_t | \mathbf{z}_{t-1}) = \mathcal{N}(\mathbf{z}_t; \sqrt{1-\beta_t} \mathbf{z}_{t-1}, \beta_t \mathbf{I}_d),  
    \label{eq:forward}
\end{equation}
where we introduced discrete time steps  $t = 1 \ldots T$, we assume $T=1000$, 
and \( \beta_t \) is a predefined noise schedule.
Note that the forward processes in \autoref{eq:forward} allows closed-form sampling at any timestep~\( t \). Using the notation \( \alpha_t := 1 - \beta_t \) and \( \bar{\alpha}_t := \prod_{s=1}^t \alpha_s \), we have:
\begin{equation}
    q(\mathbf{z}_t|\mathbf{z}_0) = \mathcal{N}(\mathbf{z}_t; \sqrt{\bar{\alpha}_t} \mathbf{z}_0, (1 - \bar{\alpha}_t)  \mathbf{I}_d).
    \label{eq:noising_step}
\end{equation}
DDPMs are a set of deep network models and sampling techniques which reverse this  process: starting from a sample
$\mathbf{z}_T \sim \mathcal{N}(\mathbf{0}, \, \mathbf{I}_d)$, one generates samples at earlier times until a sample 
$\mathbf{z}_0$  from the data distribution is obtained.

In the next section we present the particular model and 
 sampling technique we use in CLUDI. For more details on diffusion models

see recent overviews~\cite{luo2022understanding,turner2024denoising,chan2024tutorial,nakkiran2024step}. We use discrete time steps~$t$, but continuous time formulations also exist~\cite{songscore}. 
Note that our use of diffusions in continuous space to generate discrete data  resembles their use to generate discrete language tokens~\cite{dieleman2022continuous,gao2022difformer,gong2022diffuseq,li2022diffusion}.

\paragraph{Self-Supervision.}
Self-supervised learning (SSL) has emerged as a powerful paradigm for learning from unlabeled data.
The type of SSL that we employ uses a Siamese architecture~\cite{chicco2021siamese}, where  two views of the input produce different representations. The model is trained to ensure that these representations are  informative and mutually predictable. A major challenge is avoiding model collapse, where the representations become mutually predictive by minimizing the information about the inputs.  

Several mechanisms have been proposed in the SSL context to avoid collapse, such as contrastive losses~\cite{chen2020simple,jaiswal2020survey} or clustering constraints~\cite{caron2018deep,caron2021emerging}. In this work we adopt the teacher-student framework~\cite{grill2020bootstrap, chen2021exploring, caron2021emerging},
in which the teacher model generates labels for the data which the student learns to predict. As we will detail in~\Cref{sec:self-distillation}, we avoid collapse via stop-gradients, a predictor layer, and strong prior assumptions. 
We refer the reader to recent SSL 
surveys~\cite{balestriero2023cookbook,ozbulak2023know,shwartz2024compress,gui2024survey} for thorough overviews.

%% file: sections/model.tex
\section{Clustering via Diffusion} 
\label{sec:model}

Clustering via Diffusion (CLUDI) is a latent variable model for classification of the form 
\begin{align}
\hspace{-.5cm}
    p_{\theta}(k|\mathbf{x} ) = 
\int d \mathbf{z}_0 \, p(k|\mathbf{z}_0 ) 
p_{\theta}(\mathbf{z}_0|\mathbf{x} )
\simeq 
\frac{1}{B}
\sum_{i=1}^B  p(k|\mathbf{z}_0^i ) ,
\end{align}
where the last term is a Monte Carlo 
approximation obtained from $B$ samples of $p_{\theta}(\mathbf{z}_0|\mathbf{x})$,
a data-conditioned  diffusion model
that generates {\it assignment embeddings}.
The classification head is a simple logit projection followed by 
a tempered softmax, 
\begin{align}
    p(k|\mathbf{z}_0 ) &= 
    \frac{
    \exp(  \frac{[\mathbf{L}\mathbf{z}_0]_k}{\tau} ) 
    }
    {
    \sum_{j=1}^K 
    \exp(\frac{[\mathbf{L}\mathbf{z}_0]_j}{\tau}) }        
    \label{eq:L_softmax}
    \\
    & \equiv \mathbf{u}_k . 
    \label{eq:uk}
\end{align}
where $\mathbf{L} \in \mathbb{R}^{K \times d}$. 
\autoref{fig:intro_fig_2} illustrates the averaging over
$B=4$ samples of $\mathbf{z}_0$, while 
\autoref{fig:intro_fig} shows classification accuracy 
as a function of $B$ for both clean and noise-augmented data.

Starting from an initial Gaussian sample
\begin{equation}
\mathbf{z}_{T}  \sim \mathcal{N}(\mathbf{0}, F^2 \mathbf{I}_d), 
\label{eq:initial_noise}
\end{equation}
the diffusion model  outputs $\mathbf{z}_0 \in \mathbb{R}^d$. 
For  reverse sampling we adopt a stochastic version of the Denoising Diffusion Implicit Model (DDIM)~\cite{song2020denoising}, which allows sampling backwards at arbitrarily earlier times $s<t$ 
by running the  backward  dynamics 
\begin{equation}
    p_{\theta}(\mathbf{z}_{s} | \mathbf{z}_t) = \mathcal{N}(\mathbf{z}_{s}; \boldsymbol{\mu}_{\theta}(\mathbf{z}_t, \mathbf{x},s,t), F^2\sigma_{s|t}^2  \mathbf{I}_d),
    \label{eq:backwards_sample}
\end{equation}
for $s< t$, where 
\begin{align}
\boldsymbol{\mu}_{\theta}(\mathbf{z}_t, \mathbf{x}, s,t) 
&= \sqrt{\alpha_{s}} \left(\frac{\mathbf{z}_t - \sqrt{1 - \alpha_t} \, \epsilon_\theta^{(t)}(\mathbf{z}_t, \mathbf{x})}{\sqrt{\alpha_t}}\right)     
\nonumber 
    \\ 
    &\quad + \sqrt{1 - \alpha_{s} - \sigma_{s}^2} \cdot \mathbf{\epsilon}_\theta^{(t)}(\mathbf{z}_t, \mathbf{x})     
    \label{eq:ddim_step}
    \\
    \sigma_{s|t} &= \sqrt{\frac{1 - \alpha_{s}}{1 - \alpha_t} 
    \left(1-\frac{\alpha_t}{\alpha_{s}} \right)   }, 
\end{align}
and we defined
\begin{align}
\epsilon_\theta^{(t)}(\mathbf{z}_t, \mathbf{x}) &= \frac{
    \mathbf{z}_{t} - \sqrt{\bar{\alpha}_{t}} \, \tilde{\mathbf{z}}_{\theta}(\mathbf{z}_t,\mathbf{x},t) }{\sqrt{1 - {\alpha}_{t}}}.
       \label{eq:DDIM_sample}
\end{align}
Here 
$\tilde{\mathbf{z}}_{\theta}(\mathbf{z}_t,\mathbf{x},t):  \mathbb{R}^d \rightarrow \mathbb{R}^d$
is a network that predicts $\mathbf{z}_0$, and is trained 
by minimizing 
\begin{equation}
    \mathbb{E}_{\mathbf{z}_0, \mathbf{z}_t, t,\mathbf{x}} \left[ w(t) \left\lVert 
    \tilde{\mathbf{z}}_{\theta}(\mathbf{z}_t,\mathbf{x},t) 
    - \mathbf{z}_0 \right\rVert^2 \right]. 
    \label{eq:VLB}
\end{equation}
Here  $w(t)$ are fixed weights. This loss is a weighted variational lower bound on the data log-likelihood.
In modeling $\tilde{\mathbf{z}}_{\theta}$ 
we have followed~\cite{salimansprogressive}, but there exist other
possibilities, such as modeling 
$\epsilon_\theta^{(t)}$, which represents the added noise
in \autoref{eq:noising_step}.

We treat the noise 
scale $F^2$ in Eqs.(\ref{eq:initial_noise})-(\ref{eq:backwards_sample}) as an hyperparameter~\citep{gao2022difformer}. 
Note that the above backward sampling requires choosing 
the values for the time steps in \autoref{eq:backwards_sample},
a freedom we exploit below in our training scheme.

\input{images/intro_fig}

{\bf Noise schedule.} As in~\citet{li2022diffusion} we 
adopt the {\it sqrt} noise schedule with 
$\alpha_0 = \bar{\alpha}_0 =  1$ and 
\begin{align}
\bar{\alpha}_t = 1 - \sqrt{t/T+0.0001},   \,\, t \geq 1 \,.
\label{eq:sqrt_schedule2}
\end{align}
As shown  in~\autoref{fig:ss_scores}, this schedule accounts for the reduced sensitivity of the discrete labels to noise added near $t = 0$, leading to most of the noise in \autoref{eq:noising_step} being introduced at lower $t$ values. As $t$ grows, the noise addition slows down, easing the learning process of the denoising network.

\input{images/ss_scores}

\input{images/scheme}

\section{Learning via Self-Distillation}
\label{sec:self-distillation}

CLUDI follows a self-supervised learning framework based on self-distillation, similar to BYOL~\cite{grill2020bootstrap}, SimSiam~\cite{chen2021exploring} and DINO~\cite{caron2021emerging}. In this setup, two versions of the network process different representations of the same data $\mathbf{x}$, and one network predicts the output of the other. We adopt the SimSiam approach, where the teacher and student networks share weights, unlike BYOL and DINO, where the teacher is updated via an exponential moving average of the student.

A key distinction of CLUDI is that, for each data point~$\mathbf{x}$, the student has two different learning targets: the denoised 
assignment embedding~$\mathbf{z}_0$ and the classification probabilities~$u_k$ (\autoref{eq:uk}). Each target requires different strategies to avoid trivial solutions. For the embeddings $\mathbf{z}_0$,  gradients are applied only to the student while keeping the teacher fixed, and  a projection and normalization layer is added to structure the teacher’s output into a useful learning target~\cite{grill2020bootstrap, chen2021exploring, caron2021emerging}:
 These modifications prevent representational collapse and provide informative embeddings, though their precise effect remains an active area of research~\cite{tian2021understanding,wang2021towards,liu2022bridging,tao2022exploring,richemond2023edge,halvagal2023implicit}.

To avoid collapse when learning the softmax cluster probabilities $\mathbf{u}_k$, 
we enforce a uniform prior over mini-batches that regularizes the 
cross-entropy loss~\cite{amrani2022self}.

\Autoref{fig:scheme} summarizes the sequence of teacher and student operations from the input data into the loss function. In the following section, we provide more details on our choices for the teacher, the student, and the loss function.

\subsection{Teacher model}
The teacher generates cluster probabilities $\mathbf{u}$ and assignment embeddings, $\mathbf{z}_0$ which serve as training targets for the student. It receives $N$ input feature vectors $\mathbf{x}^i \in\mathbb{R}^n$, $i \in [1,N]$, and runs the denoising algorithm $B$ times for each of them, yielding $B \times N$ 
denoised embeddings $\tilde{\mathbf{z}}_0^{b,i} \in \mathbb{R}^{d}$. The time schedule of the teacher denoising is chosen to 
contain 25 equally-spaced timesteps from $t = T=1000$ to $t=0$, offering a balance between efficient denoising and capturing essential cluster characteristics.
Acting on $\tilde{\mathbf{z}}_0^{b,i}$
with $\mathbf{L}$ and a softmax (see 
\autoref{eq:L_softmax}-\autoref{eq:uk}), 
yields $B \times N$  probability targets $\mathbf{u}^{b,i} \in \mathbb{R}^{ K}$ for the student. 

Empirically, however, 
the learning is less effective when the student's targets directly the denoised embeddings~$\tilde{\mathbf{z}}_0^{b,i}$. 
Instead, we map  the probabilities~$\mathbf{u}^{b,i}$
back to the embedding space 
by means of  
an embedding 
matrix~$\mathbf{E}\in\mathbb{R}^{d \times K}$, and then normalize and scale the
target embeddings as 
\begin{align}
    \mathbf{z}_0^{b,i}  & = \sqrt{d} \frac{\mathbf{E}\mathbf{u}^{b,i}}
    {\| \mathbf{E}\mathbf{u}^{b,i}\|_2} .
    \label{eq:denoised_target}
\end{align}

The role of $\mathbf{E}$ is akin to the predictor network in BYOL or SimSiam,
which assists in generating effective representations during training. 
Once the model is fully trained, however,~$\mathbf{E}$~is no longer required. 
Given the above  embeddings, the teacher class  probabilities are 
\begin{equation}
\mathbf{u}^{b,i} =    \frac{
    \exp(  \frac{[\mathbf{L}\mathbf{z}_0^{b,i}]_k}{\tau} ) 
    }
    {
    \sum_{j=1}^K 
\exp(\frac{[\mathbf{L}\mathbf{z}_0^{b,i}]_j}{\tau}) }    \,.     
\label{eq:teacher_probs}
\end{equation}
\input{images/tsne}

\subsection{Student model}
The student network aims to predict both
targets in \autoref{eq:denoised_target}
and \autoref{eq:teacher_probs}. 
 
Since the input $\mathbf{x}$ consists of pre-trained features, 
we use an abstract augmentation strategy.  
For each data feature vector $\mathbf{x}^i \in \mathbb{R}^D$, $i \in [1,N]$, we create $B$ augmented versions $\mathbf{x}^{b,i}$,     
$b \in [1,B]$, each  obtained by first zeroing the components of $\mathbf{x}^i$ with probability~$0.2$, and then adding zero-mean Gaussian noise, with variance~$\sigma^2$ sampled uniformly in  $[0.1,0.3]$.  We present the student with noisy versions of the 
teacher’s outputs in \autoref{eq:denoised_target}, 
\begin{equation} 
\mathbf{z}_{t_b}^{b,i} \sim \mathcal{N}\left(\sqrt{\bar{\alpha}_{t_b}} \mathbf{z}_0^{b,i}, (1 - \bar{\alpha}_{t_b}) F^2 \mathbf{I_d}\right),
\label{eq:noise_sample} 
\end{equation}
where $t_b$ is sampled uniformly from $[0,T]$.
We denote the student predictions for the denoised assignment embedding and class probabilities as 
\begin{align}
\hat{\mathbf{z}}_0^{b,i} &=     \tilde{\mathbf{z}}_{\theta}(\mathbf{z}_{t_b}^{b,i},\mathbf{x}^{b,i},t_b) \,,
\label{eq:one_tstep_denoise}
\\
\mathbf{\hat{u}}^{b,i} &=    \frac{
    \exp(  \frac{[\mathbf{L}\hat{\mathbf{z}}_0^{b,i}]_k}{\tau} ) 
    }
    {
    \sum_{j=1}^K 
\exp(\frac{[\mathbf{L}\hat{\mathbf{z}}_0^{b,i}]_j}{\tau}) }    \,.   
\label{eq:student_probs}
\end{align}

\subsection{Augmented views}
Having presented the teacher and student models, we
note that their different views of the data
and  assignment embeddings 
originate from (i) the stochastic nature 
of the teacher denoising, which starts 
with pure noise in \autoref{eq:initial_noise}
and performs 25 sampling steps,  (ii) the 
augmentation of the student feature vector (feature dropout + Gaussian noise) and
(iii) the noise added in \autoref{eq:noise_sample} to the embedding that the student is required to denoise. We remark that unlike common augmentation strategies acting on raw images (cropping, color alterations, etc), our augmentations act directly on the data features or their assignment embeddings. 
In particular, the views generated by 
(i) and (iii) exploit the intrinsic randomness of the diffusion model.  
\subsection{Loss functions} 

\paragraph{Embeddings.} 
We employ an MSE loss between 
the teacher denoised embeddings~$\mathbf{z}_0^{b,i}$ from~\autoref{eq:denoised_target}
and the student-predicted embeddings $\hat{\mathbf{z}}_0^{b,i}$
from~\autoref{eq:one_tstep_denoise}:
\begin{equation}
    \ell_{dif}(\mathbf{z}_0^{b,i}, \hat{\mathbf{z}}_0^{b,i}) = \|\mathbf{z}_0^{b,i} - \hat{\mathbf{z}}_0^{b,i}\|^2. 
\end{equation}

\paragraph{Class  probabilities.} 
We use here a more explicit notation for the teacher and student class probabilities, respectively, as defined 
\autoref{eq:teacher_probs} and \autoref{eq:student_probs}, 
\begin{align}
    p(k|\mathbf{z}_0^{b,i}) = \mathbf{u}_k^{b,i},
    \qquad \quad 
    p(k|\hat{\mathbf{z}}_0^{b,i}) = \mathbf{\hat{u}}_k^{b,i} \,.
\end{align}
Naively using a cross entropy loss, 
\begin{align}
\ell(\hat{\mathbf{z}}_0^{b,i}, \mathbf{z}_0^{b,i}) = -\sum_k
p(k|\mathbf{z}_0^{b,i}) 
\log p(k|\hat{\mathbf{z}}_0^{b,i}), 
\label{eq:cross_entropy_loss}
\end{align}
quickly degenerates to a solution that puts all the probability on a single category. 
We regularize this loss using an idea from~\citet{amrani2022self}, which amounts to treating the indices $(b,i)$
as random variables with a uniform prior
$p(\hat{\mathbf{z}}_0^{b,i}) = 1/(NB)$ 
and a distribution conditioned on $k$ given by a column softmax  
\begin{align}
    p(\mathbf{z}_0^{b,i} |k) &= 
    \frac{
    \exp(  \frac{[\mathbf{L}\mathbf{z}_0^{b,i}]_k}{\tau_{col}} ) 
    }
    {
    \sum_{b',i'=1}^{B,N}     \exp(\frac{[\mathbf{L}\mathbf{z}_0^{b',i'}]_k}{\tau_{col}}) } \,,        
    \label{eq:column_softmax}
\end{align}
with its own temperature $\tau_{col}$.  
Assuming also a uniform class prior $p(k)=1/K$, 
Bayes rule and the law of total probability imply 
\begin{align}
p(k|\mathbf{z}_0^{b,i}) 
& = \frac{p(\mathbf{z}_0^{b,i}|k)p(k) }{p(\mathbf{z}_0^{b,i})} 
= 
\frac{p(\mathbf{z}_0^{b,i}|k) }{\sum_{k=1}^K p(\mathbf{z}_0^{b,i}|k)} \,,
\\
p(k|\hat{\mathbf{z}}_0^{b,i})  & = 
\frac{p(k) p(k|\hat{\mathbf{z}}_0^{b,i})}{p(k)}
= \frac{(NB/K)p(k|\hat{\mathbf{z}}_0^{b,i})}{\sum_{b',i'=1}^{B,N}
p(k|\hat{\mathbf{z}}_0^{b',i'}) 
}\,.
\end{align}
Inserting these expressions in \autoref{eq:cross_entropy_loss} leads to a loss which was shown in~\cite{amrani2022self} not to admit collapsed distributions as optimal solutions.
In practice, we use a symmetric 
version of~\autoref{eq:cross_entropy_loss},  
\begin{align}
\ell_{cls}(b,i) = \frac12   
\left( 
\ell(\hat{\mathbf{z}}_0^{b,i}, \mathbf{z}_0^{b,i})  +
\ell(\mathbf{z}_0^{b,i}, \hat{\mathbf{z}}_0^{b,i})  
\right) .
\label{eq:class_loss}
\end{align}

\paragraph{Weighting the minibatch elements.} 
Addressing the varying difficulty of the predictions
of $\mathbf{z}_0^{b,i}, {\mathbf{u}}^{b,i}$ based on the sampled timestep $t_b$, we incorporate  
the Min-SNR-\(\gamma\)~\cite{Hang_2023_ICCV} loss weighting strategy, which treats each timestep's denoising task as distinct and assigns weights based on their difficulty:
\begin{equation}
    \text{SNR}_{t_b} = \frac{\bar{\alpha}_{t_b}}{1 - \bar{\alpha}_{t_b}},
\end{equation}
\begin{equation}
    w_{b} = \frac{\max(\text{SNR}_{t_b}, \gamma)}{\text{SNR}_{t_b} + 1},
\end{equation}
where \(\gamma\) is a predefined threshold set to 5, enhancing the stability and ensuring that no single noise level dominates during training. The overall loss function, which incorporates these weights, is structured as follows:
\begin{equation}
    \mathcal{L} = \frac{1}{BN} \sum_{b,i=1}^{B,N} w_b \left( \ell_{dif}(\mathbf{z}_0^{b,i}, \hat{\mathbf{z}}_0^{b,i}) + \lambda 
    \ell_{cls}(b,i) 
    \right). 
    \label{eq:final_loss}
\end{equation}
This formulation enables a nuanced control over the learning process, adapting to the challenges posed by different levels of noise.

%% file: images/intro_fig.tex
\begin{figure}[!t]
    \centering
\includegraphics[width=\linewidth]{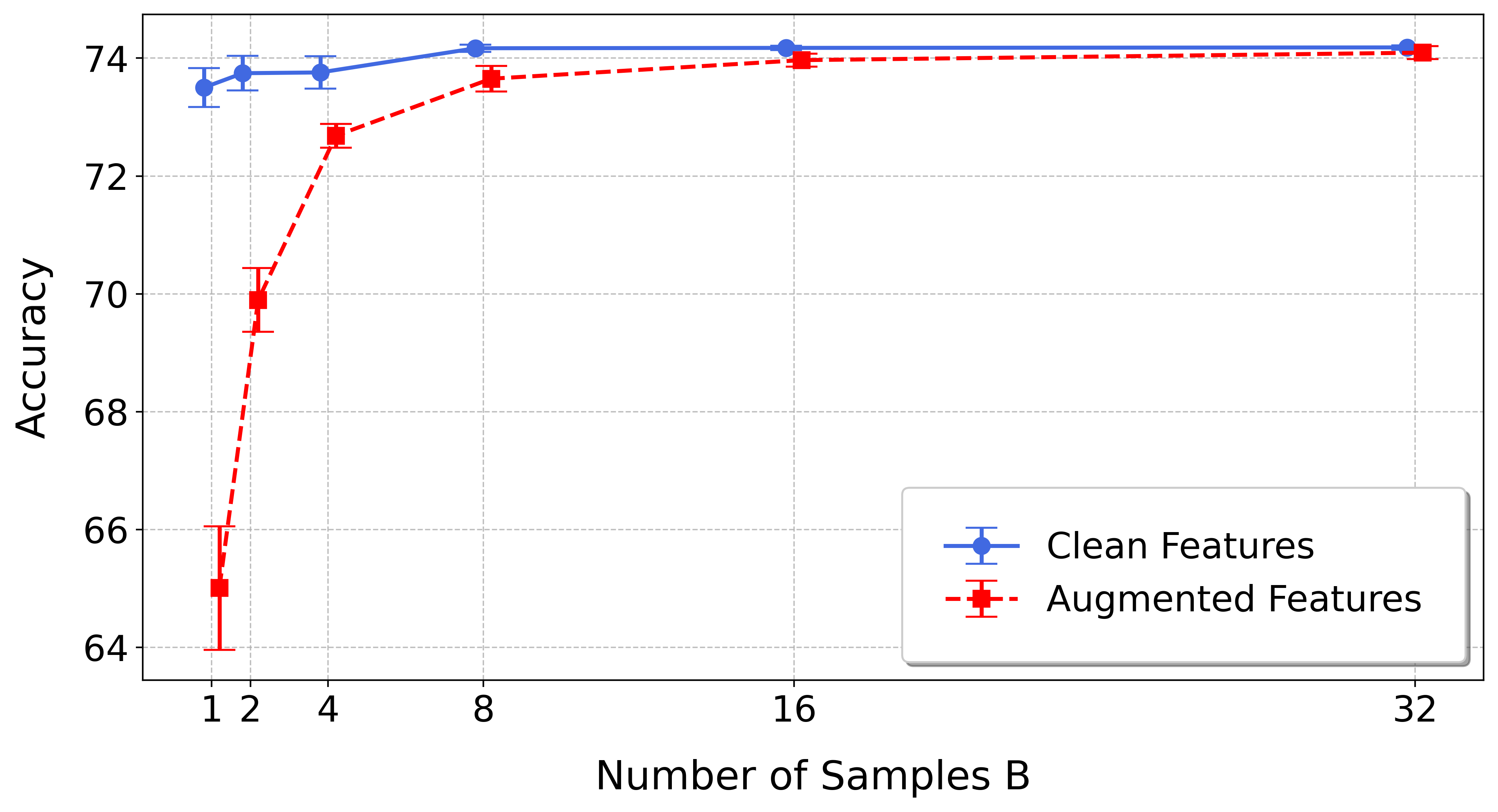}
\caption{{\bf Classification accuracy for clean and augmented inputs.}  The model classification accuracy on augmented data (feature dropout plus Gaussian noise)  becomes similar to that of clean data as the number of data samples grows. Results from ImageNet 100 validation data. Standard deviation based on 10 repetitions.}

    \label{fig:intro_fig}
\end{figure}

%% file: images/ss_scores.tex
\begin{figure}[!h]
    \centering
\includegraphics[width=\linewidth]
{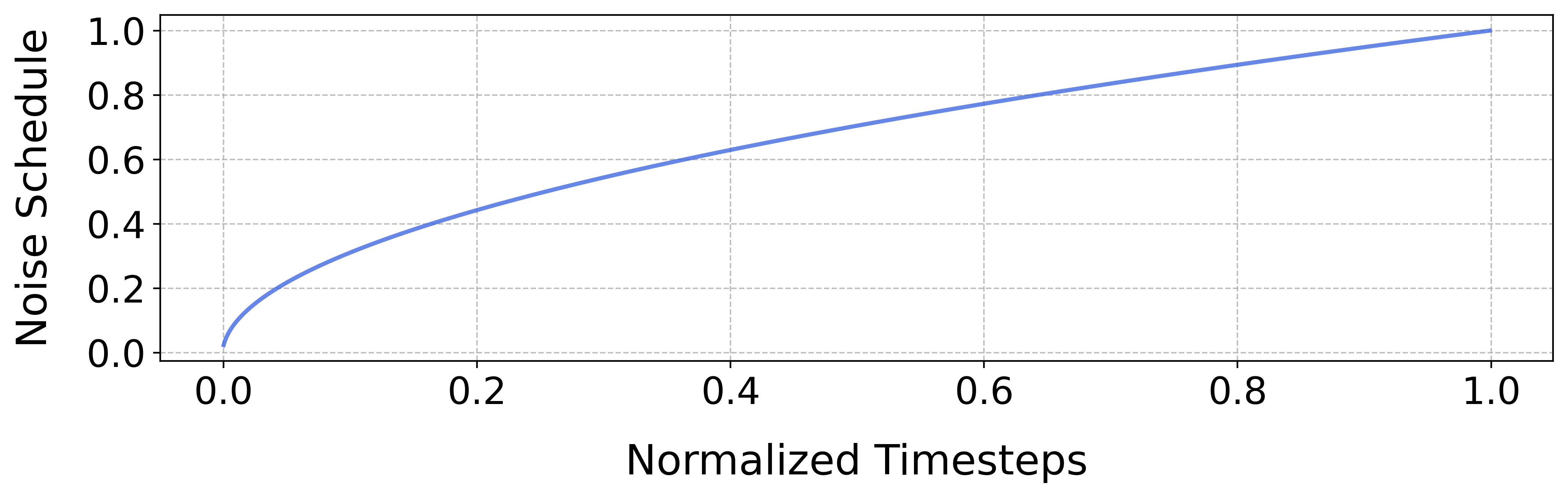}
\caption{{\bf \textit{sqrt} noise scheduling.} 
The noise grows faster
near \( t = 0 \), reflecting reduced sensitivity to noise at early timesteps, and grows gradually slower at later times.}

    \label{fig:ss_scores}
\end{figure}

%% file: images/scheme.tex
\begin{figure*}[h]
\centering
\includegraphics[width=\textwidth]{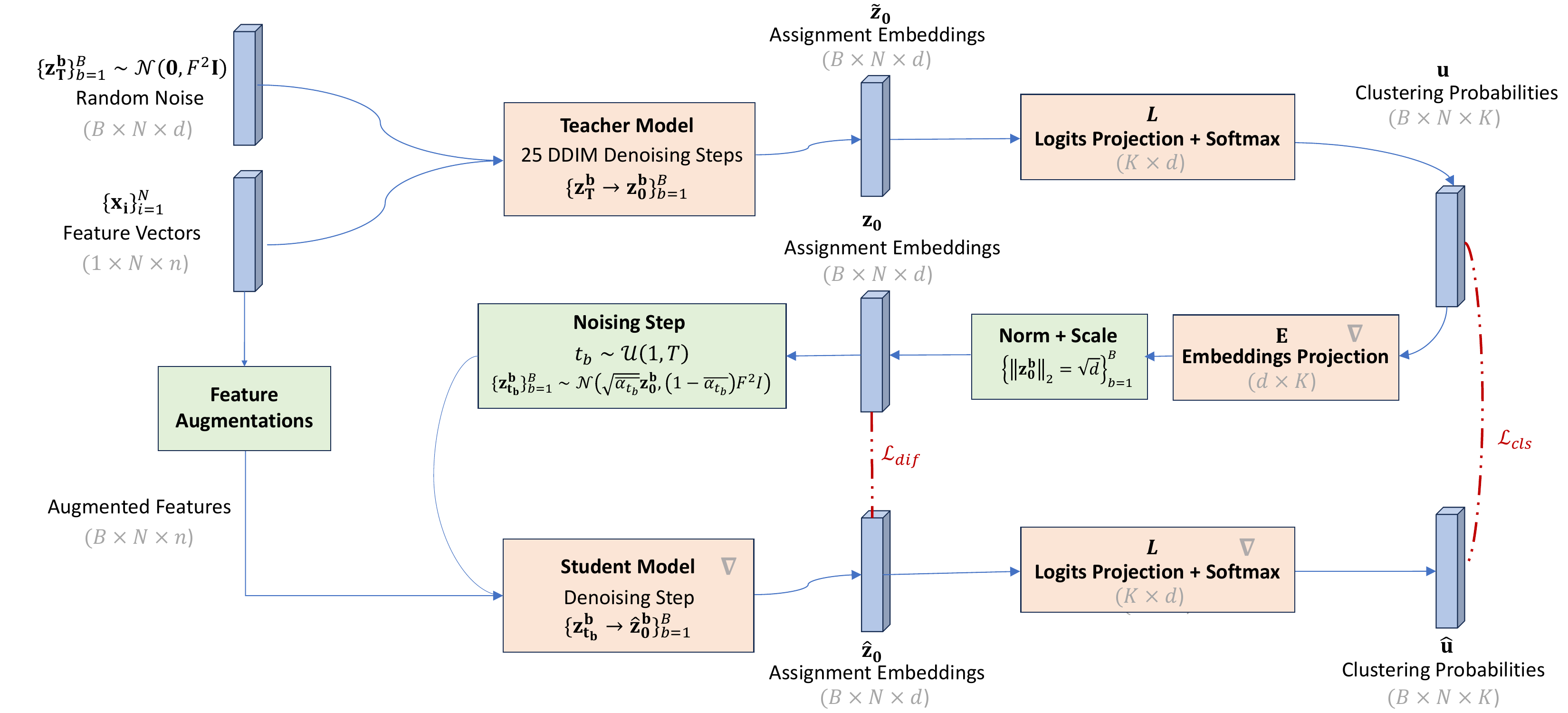}
\caption{{\bf Overview of CLUDI’s training phase.} 
Given a set of image features $\mathbf{x}$, the teacher model generates denoised assignment embeddings $\tilde{\mathbf{z}}_0$ which are used to create two targets for the student: (i) clustering probabilities $\mathbf{u}$ and (ii) assignment embeddings $\mathbf{z}_0$,  obtained from $\mathbf{u}$ via a 
predictor layer. The student network aims to predict both targets based on a version of 
$\mathbf{x}$ corrupted by feature dropout plus Gaussian noise.  Components whose parameters are updated via gradient-based optimization are marked with the $\nabla$ symbol.}
\label{fig:scheme}
\end{figure*}

%% file: images/tsne.tex
\begin{figure*}[h!]
\includegraphics[width=\textwidth,height=0.42\textheight]{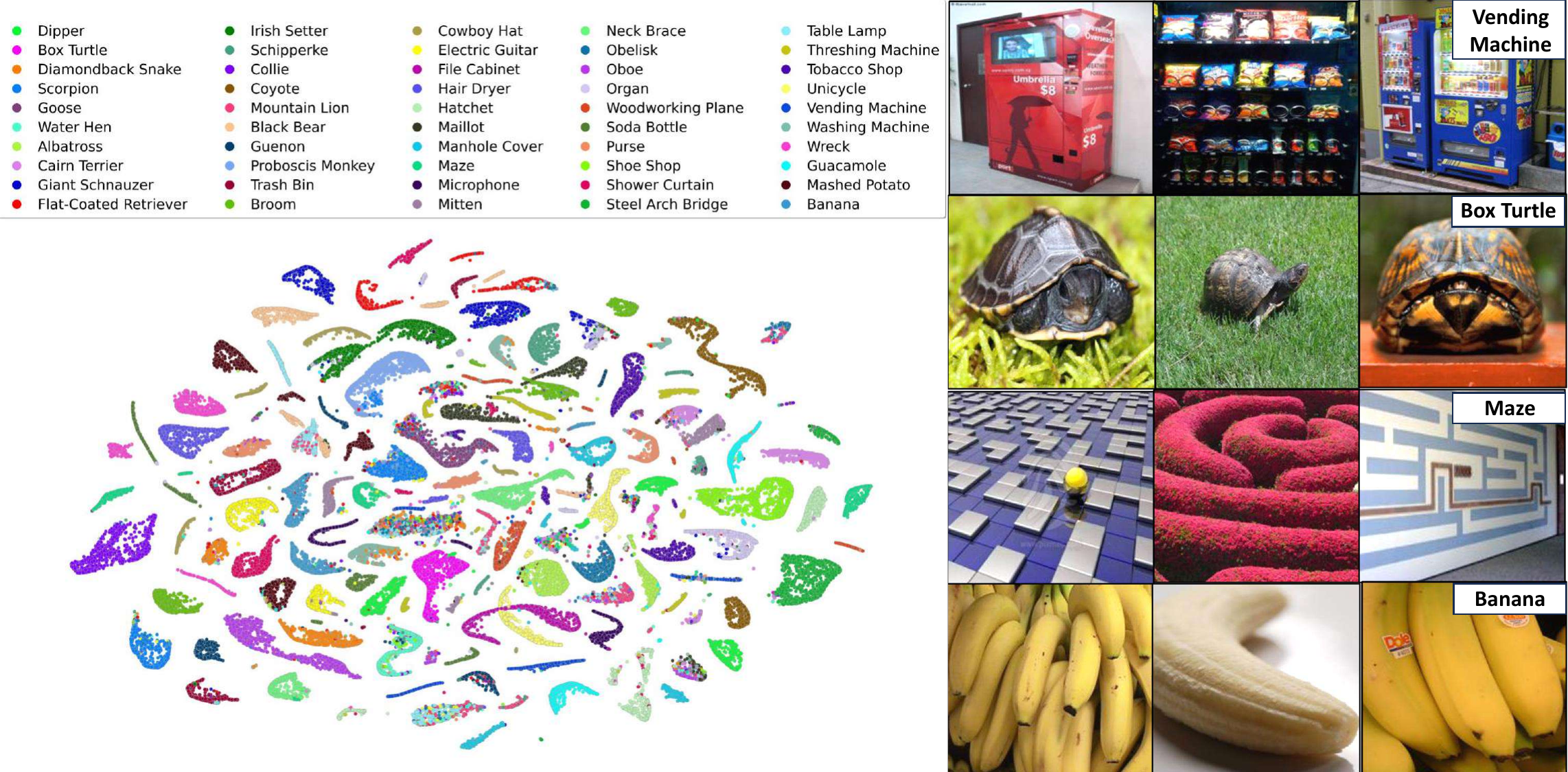}
\caption{{\it Left:} t-SNE visualization of the assignment embedding space of ImageNet 50  demonstrating the model's ability to organize data points into well-separated clusters. {\it Right:} Examples of correctly classified images}
    \label{fig:tsne_plot}
\end{figure*}

%% file: sections/experiments.tex
\input{tables/small_datasets}
\input{tables/imagenet_subsets}

\section{Experiments}
\label{sec:experiments}

\textbf{Datasets.} We evaluate CLUDI on a comprehensive suite of benchmark datasets to rigorously assess its scalability, adaptability, and clustering performance. The datasets include subsets of ImageNet \cite{deng2009imagenet}, Oxford-IIIT Pets \cite{parkhi2012cats} (with $K=32$), Oxford 102 Flower \cite{nilsback2008automated}, Caltech 101 \cite{fei2004learning}, CIFAR-10 \cite{krizhevsky2009learning}, and STL-10 \cite{coates2011analysis}. Each dataset introduces unique challenges: the ImageNet subsets cover broad and diverse categories, providing a robust test of scalability (see \autoref{table:imagenet-subsets}); Oxford-IIIT Pets and Oxford 102 Flower focus on fine-grained distinctions that assess CLUDI's precision; Caltech 101 evaluates generalization across diverse object types; and CIFAR-10 and STL-10 offer additional complex data that further validate CLUDI's ability to handle intricate clustering tasks (\autoref{table:small_datasets}).

\textbf{Evaluation Metrics.} We use three popular  metrics~\cite{fahad:2014:NMI_ARI}:
(i) \textit{Normalized Mutual Information (NMI)} quantifies shared information between predicted and ground-truth clusters; (ii) \textit{Clustering Accuracy (ACC)} measures the alignment of predictions with true labels;
(iii) \textit{Adjusted Rand Index (ARI)} adjusts for chance, providing a robust measure of similarity between predicted and true clusters.

\textbf{Experimental Setup.} For all experiments, we use features from DINO~\cite{caron2021emerging} based on Vision Transformers (ViT-S/16 and ViT-B/16) pre-trained on ImageNet. We present comparisons with four leading self-supervised clustering models: SCAN~\cite{van2020scan}, Propos~\cite{huang2022learning}, TSP~\cite{zhou2022deep} and TEMI~\cite{adaloglou2023exploring}.
We implement a variant of Self-Classifier~\cite{amrani2022self}, denoted as \textit{Self-Classifier*}, in which the feature extractor is frozen and only the classification heads are optimized.
This setup ensures that CLUDI and Self-Classifier* share identical feature representations, enabling a more direct and fair comparison.

All CLUDI results were obtained by running the embedding denoising for 100 equally spaced time steps. 
\Autoref{fig:acc_lambda} presents the ablation of the classification-loss weight~\(\lambda\), 
while \Autoref{fig:max_accuracy_combined} illustrates the results of scanning different values of the embedding dimension \(d\) in \autoref{eq:final_loss}.
For additional implementation details and ablations, see our Appendix.

\textbf{Results.} CLUDI achieves state-of-the-art performance across all tested datasets, consistently outperforming previous approaches on clustering tasks of varying complexity. As detailed in 
\autoref{table:small_datasets}
and~\autoref{table:imagenet-subsets}, CLUDI significantly surpasses established baselines in NMI, ACC, and ARI, especially on ImageNet subsets, showcasing its robustness in both general and fine-grained clustering scenarios.

\textbf{Qualitative Analysis.} A t-SNE plot (\autoref{fig:tsne_plot}) of CLUDI's embeddings on ImageNet-50 demonstrates its  ability to form well-separated clusters, reinforcing its quantitative metrics and illustrating its effectiveness in organizing complex data structures into distinct clusters. This visualization highlights CLUDI’s capacity to capture subtle 
inter-class variations.%

\input{images/ablations_main}
\textbf{Limitations.}
The effectiveness of the CLUDI model is influenced by the choice of the diffusion parameter \( F^2 \) and the embedding dimensionality \( d \), both of which play critical roles in determining clustering quality. While these parameters were tuned to optimize performance on the tested datasets, further refinement may be necessary for new data distributions or specific clustering tasks. Additionally, although CLUDI demonstrates strong capability in generating well-separated clusters, its performance can be impacted when scaling to a large number of clusters, as maintaining high-quality, distinct embeddings becomes increasingly complex with higher cluster counts. 

\section{Conclusion} 
\label{sec:conclusion}
In this work we introduced a novel use of diffusion models to generate clustering embeddings of pretrained data features. Our experimental results underscore CLUDI’s advantages over both traditional and contemporary clustering techniques, validating its robustness, flexibility, and superior self-supervised clustering performance across diverse datasets and visual challenges. Future studies could build upon this work by investigating adaptive or data-driven hyperparameter selection techniques, as well as advanced clustering frameworks, such as hierarchical or multi-scale methods, potentially more scalable to large $K$ settings.

%% file: tables/small_datasets.tex
\begin{table}[t!]
\caption{\textbf{Clustering performances on smaller datasets.} 
The results for SCAN, Propos, TSP, and TEMI on CIFAR-10 and STL-10 are from  the original papers, except for TEMI (ViT-S/16), which we trained using the official code. 
For the Oxford and Caltech datasets, we trained all models. 
The best result is shown in \textbf{bold}, the second best is \underline{underlined}.}
\centering
\resizebox{1\linewidth}{!}{ 
\begin{tabular}{lccc}
    \toprule
    \textbf{Methods} & \textbf{NMI (\%)} & \textbf{ACC (\%)} & \textbf{ARI (\%)} \\
      \midrule
    \multicolumn{4}{c}
    {\textbf{CIFAR 10}} \\
    SCAN (Resnet50) & 79.7 & 88.3 & 77.2 \\
    Propos (Resnet18)& 88.6 & 94.3 & 88.4 \\
    TSP (ViT-S/16) & 84.7 & 92.1 & 83.8\\
    TSP (ViT-B/16) & 88.0 & 94.0 & 87.5\\
    TEMI (ViT-S/16) & 85.4 & 92.7 & 84.8 \\
    TEMI (ViT-B/16) & \underline{88.6} & \underline{94.5} & \underline{88.5} \\
    Self-Classifier* (ViT-S/16) & 83.8 & 91.2 & 82.1 \\
    Self-Classifier* (ViT-B/16)  & 84.2 & 89.0 & 80.3  \\
    \textbf{Ours (ViT-S/16)} &  88.0 & 94.2 & 87.7  \\
    \textbf{Ours (ViT-B/16)}  & \textbf{89.6} & \textbf{95.3} & \textbf{89.8}  \\
          \midrule
    \multicolumn{4}{c}
    {\textbf{STL 10}} \\
    SCAN (Resnet50) & 69.8 & 80.9 & 64.6 \\
    Propos (Resnet18)& 75.8& 86.7 & 73.7 \\
    TSP (ViT-S/16) & 94.1 & 97.0 & 93.8\\
    TSP (ViT-B/16) & 95.8 & 97.9 & 95.6\\
    TEMI (ViT-S/16) & 85.0 & 88.8 & 80.1 \\
    TEMI (ViT-B/16) & \underline{96.5} &  \underline{98.5} & \underline{96.8} \\
    Self-Classifier* (ViT-S/16) & 90.5 & 83.1 & 82.4 \\
    Self-Classifier* (ViT-B/16)  & 91.5 & 87.7  &85.7  \\
    \textbf{Ours (ViT-S/16)} &  95.7 & 98.2 & 96.1  \\
    \textbf{Ours (ViT-B/16)}  & \textbf{96.8 }& \textbf{98.7}  & \textbf{97.1} \\

    \midrule
    \multicolumn{4}{c}{\textbf{Oxford-IIIT Pets}} \\
    TEMI (ViT-S/16) & 69.7 & 49.3 & 41.0 \\
    TEMI (ViT-B/16) & 71.1 & 47.0 & 41.7 \\
    Self-Classifier* (ViT-S/16) & 82.7 & 67.5 & 59.2 \\
    Self-Classifier* (ViT-B/16) & 83.5 & 68.2 & 63.0 \\
    \textbf{Ours (ViT-S/16)} & \textbf{87.3} & \textbf{74.1} & \textbf{71.6} \\
    \textbf{Ours (ViT-B/16)} & \underline{86.7} & \underline{73.8} & \underline{71.1} \\
    \midrule
    \multicolumn{4}{c}
    {\textbf{Oxford 102 Flower}} \\
    TEMI (ViT-S/16) & 50.1 & 26.0 & 14.2 \\
    TEMI (ViT-B/16) & 50.2 & 25.9 & 16.9 \\
     Self-Classifier* (ViT-S/16) & 69.1 & 51.5 & 35.4 \\
    Self-Classifier* (ViT-B/16) & 72.5 & 57.8 & 42.9 \\
    \textbf{Ours (ViT-S/16)} & \underline{76.1} & \underline{62.2} & \underline{52.6} \\
    \textbf{Ours (ViT-B/16)} & \textbf{81.5} & \textbf{69.7} & \textbf{61.8} \\
    \midrule
    \multicolumn{4}{c}{\textbf{Caltech 101}} \\
    TEMI (ViT-S/16) & 78.9 & 50.2 & 35.6 \\
    TEMI (ViT-B/16) & 80.4 & 51.4 & 36.9 \\
    Self-Classifier* (ViT-S/16) & 82.5 & 56.1 & 59.4 \\
    Self-Classifier* (ViT-B/16) & 83.5 & 58.2 & 61.2 \\
    \textbf{Ours (ViT-S/16)} & \underline{86.5} & \underline{66.7} & \underline{65.7} \\
    \textbf{Ours (ViT-B/16)} & \textbf{87.9} & \textbf{68.1} & \textbf{66.3} \\
    \bottomrule
\end{tabular}
}
\label{table:small_datasets}
\end{table}

%% file: tables/imagenet_subsets.tex
\begin{table}[t!]
\caption{\textbf{Clustering performances on ImageNet subsets.}
The results for SCAN, Propos and TEMI are from 
the original papers. 
The best result is shown in \textbf{bold}, the second best is \underline{underlined}. }
\centering
\resizebox{1\linewidth}{!}{ 
\begin{tabular}{lccc}
    \toprule
    \textbf{Methods} & \textbf{NMI (\%)} & \textbf{ACC (\%)} & \textbf{ARI (\%)} \\
    \midrule
    \multicolumn{4}{c}{\textbf{ImageNet 50}} \\
    SCAN (Resnet50) & 82.2 & 76.8 & 66.1 \\
    Propos (Resnet50) & 82.8 & - & 69.1 \\
    TEMI (ViT-S/16) & 84.2 & 77.8 & 68.4 \\
    TEMI (ViT-B/16) & 86.1 & 80.1 & 71.0 \\
    Self-Classifier* (ViT-S/16) & 87.1 & 76.3 & 71.1 \\
    Self-Classifier* (ViT-B/16) & 87.8 & 77.1 & 73.2 \\
    \textbf{Ours (ViT-S/16)} & \underline{90.1} & \underline{81.3} & \underline{75.8} \\
    \textbf{Ours (ViT-B/16)} & \textbf{91.2} & \textbf{82.1} & \textbf{76.2} \\
    \midrule
    \multicolumn{4}{c}{\textbf{ImageNet 100}} \\
    SCAN (Resnet50) & 80.8 & 68.9 & 57.6 \\
    Propos (Resnet50) & 83.5 & - & 63.5 \\
    TEMI (ViT-S/16) & 83.3 & 72.5 & 62.3 \\
    TEMI (ViT-B/16) & 85.6 & \underline{75.0} & 65.4 \\
    Self-Classifier* (ViT-S/16) & 83.9 & 68.9 & 61.8 \\
    Self-Classifier* (ViT-B/16) & 86.2 & 71.4 & 64.3 \\
    \textbf{Ours (ViT-S/16)} & \underline{86.5} & 74.3 & \underline{67.5} \\
    \textbf{Ours (ViT-B/16)} & \textbf{87.1} & \textbf{76.6} & \textbf{67.8} \\
    \midrule
    \multicolumn{4}{c}{\textbf{ImageNet 200}} \\
    SCAN (Resnet50) & 77.2 & 58.1 & 47.0 \\
    Propos (Resnet50) & 80.6 & - & 53.8 \\
    TEMI (ViT-S/16) & 82.7 & 71.9 & 59.8 \\
    TEMI (ViT-B/16) & 85.2 & 73.1 & \underline{62.1} \\
    Self-Classifier* (ViT-S/16) & 80.5 & 54.5 & 46.7 \\
    Self-Classifier* (ViT-B/16) & 78.3 & 53.1 & 44.1 \\
    \textbf{Ours (ViT-S/16)} & \underline{85.9} & \underline{73.2} & 60.9 \\
    \textbf{Ours (ViT-B/16)} & \textbf{86.1} & \textbf{73.7} & \textbf{63.2} \\
    \bottomrule
\end{tabular}}
\label{table:imagenet-subsets}
\end{table}

%% file: images/ablations_main.tex


\begin{figure}[!b]
  \begin{center}  
  \includegraphics[width=0.9\columnwidth]{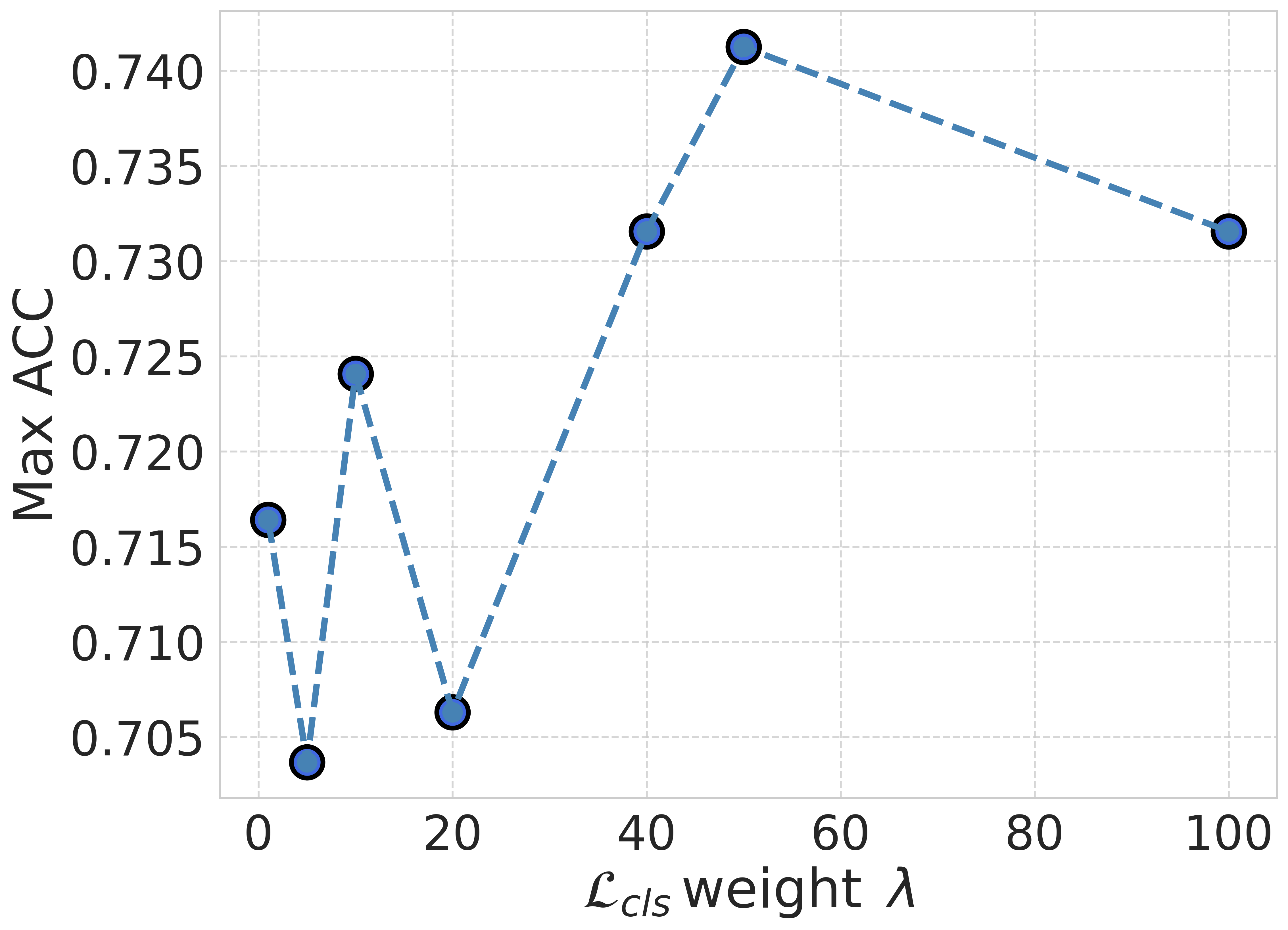}
  \caption{\textbf{Ablation Study on the $\mathcal{L}_{cls}$ weight $\lambda$.}
    Each point in the curves shows the maximum validation accuracy on ImageNet-100 achieved during training.}
    \label{fig:acc_lambda}
    \end{center}
\end{figure}

\begin{figure}[!h]
  \centering
  \includegraphics[width=0.9\columnwidth]{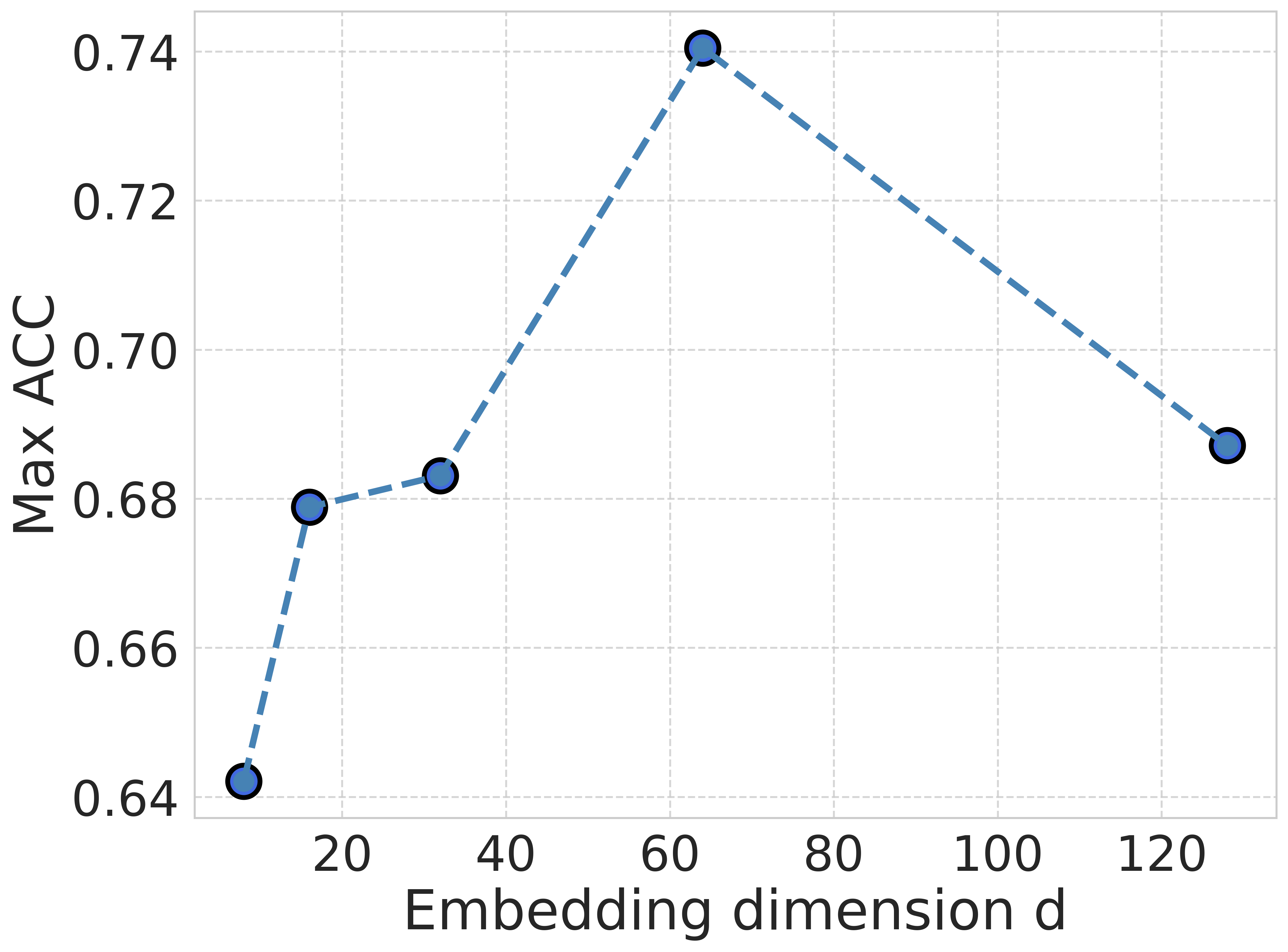}
  \caption{\textbf{Embedding dimension selection $d$.}
    Each point in the curves shows the maximum validation accuracy on ImageNet-100 achieved during training.}
  \label{fig:max_accuracy_combined}
\end{figure}

%% file: images/rescaling_factor_x.tex
\newcommand{\displayRescalingImages}[1]{
    \begin{figure*}[h!t]
        \centering
        \begin{subfigure}{0.31\textwidth}
            \centering
            \includegraphics[width=\textwidth]{images/rescaling_factor_#1/max_NMI_rescaling_#1.png}
            \caption{}
        \end{subfigure}%
        \hspace{0.5em}
        \begin{subfigure}{0.31\textwidth}
            \centering
            \includegraphics[width=\textwidth]{images/rescaling_factor_#1/max_ACC_rescaling_#1.png}
            \caption{}
        \end{subfigure}
        \hspace{0.5em}
        \begin{subfigure}{0.31\textwidth}
            \centering
            \includegraphics[width=\textwidth]{images/rescaling_factor_#1/max_ARI_rescaling_#1.png}
            \caption{}
        \end{subfigure}
        \caption{Max NMI (a), ACC (b), and ARI (c) across embedding dimensions for rescaling factor \( F^2 = #1 \).}
        \label{fig:rescaling_images_F2_#1}
    \end{figure*}
}

\newcommand{\displayThreeImagesWithCaption}[2]{
    \begin{figure*}[h!t]
        \centering
        \begin{subfigure}{0.31\textwidth}
            \centering
            \includegraphics[width=\textwidth]{images/rescaling_factor_#1/max_NMI_rescaling_#1.png}
            \caption{}
        \end{subfigure}%
        \hspace{0.5em}
        \begin{subfigure}{0.31\textwidth}
            \centering
            \includegraphics[width=\textwidth]{images/rescaling_factor_#1/max_ACC_rescaling_#1.png}
            \caption{}
        \end{subfigure}
        \hspace{0.5em}
        \begin{subfigure}{0.31\textwidth}
            \centering
            \includegraphics[width=\textwidth]{images/rescaling_factor_#1/max_ARI_rescaling_#1.png}
            \caption{}
        \end{subfigure}
        \caption{#2} 
        \label{fig:rescaling_images_F2_#1}
    \end{figure*}
}

\newcommand{\displayFullPageImages}[3]{
    \begin{minipage}{0.31\textwidth}
        \centering
        \includegraphics[width=\textwidth]{images/rescaling_factor_#1/max_NMI_rescaling_#1.png}
        \par\vspace{0.5em}NMI for \( F^2 = #1 \)
    \end{minipage}%
    \hspace{0.5em}
    \begin{minipage}{0.31\textwidth}
        \centering
        \includegraphics[width=\textwidth]{images/rescaling_factor_#1/max_ACC_rescaling_#1.png}
        \par\vspace{0.5em}ACC for \( F^2 = #1 \)
    \end{minipage}%
    \hspace{0.5em}
    \begin{minipage}{0.31\textwidth}
        \centering
        \includegraphics[width=\textwidth]{images/rescaling_factor_#1/max_ARI_rescaling_#1.png}
        \par\vspace{0.5em}ARI for \( F^2 = #1 \)
    \end{minipage}

    \vspace{1em} 
    \begin{minipage}{0.31\textwidth}
        \centering
        \includegraphics[width=\textwidth]{images/rescaling_factor_#2/max_NMI_rescaling_#2.png}
        \par\vspace{0.5em}NMI for \( F^2 = #2 \)
    \end{minipage}%
    \hspace{0.5em}
    \begin{minipage}{0.31\textwidth}
        \centering
        \includegraphics[width=\textwidth]{images/rescaling_factor_#2/max_ACC_rescaling_#2.png}
        \par\vspace{0.5em}ACC for \( F^2 = #2 \)
    \end{minipage}%
    \hspace{0.5em}
    \begin{minipage}{0.31\textwidth}
        \centering
        \includegraphics[width=\textwidth]{images/rescaling_factor_#2/max_ARI_rescaling_#2.png}
        \par\vspace{0.5em}ARI for \( F^2 = #2 \)
    \end{minipage}

    \vspace{1em} 
    \begin{minipage}{0.31\textwidth}
        \centering
        \includegraphics[width=\textwidth]{images/rescaling_factor_#3/max_NMI_rescaling_#3.png}
        \par\vspace{0.5em}NMI for \( F^2 = #3 \)
    \end{minipage}%
    \hspace{0.5em}
    \begin{minipage}{0.31\textwidth}
        \centering
        \includegraphics[width=\textwidth]{images/rescaling_factor_#3/max_ACC_rescaling_#3.png}
        \par\vspace{0.5em}ACC for \( F^2 = #3 \)
    \end{minipage}%
    \hspace{0.5em}
    \begin{minipage}{0.31\textwidth}
        \centering
        \includegraphics[width=\textwidth]{images/rescaling_factor_#3/max_ARI_rescaling_#3.png}
        \par\vspace{0.5em}ARI for \( F^2 = #3 \)
    \end{minipage}
}

%% file: sections/supplementary.tex
\clearpage
\appendix
\setcounter{figure}{0}    
\renewcommand\thefigure{S\arabic{figure}}

In this Appendix we present  the results of several hyperparameter ablations. All the curves shown correspond to performance metrics evaluated on the validation set of ImageNet 100. We also show some clustering examples on ImageNet 50.  

\section{Hyperparameters}
The model requires three hyperparameters: the embedding dimension $d$, the noise rescaling factor $F^2$ in~\autoref{eq:initial_noise} and the coefficient 
 \(\lambda\)  on the loss
 in \autoref{eq:final_loss}. 
A systematic scan on the validation set of ImageNet 100 yielded the optimal values $d=64$, $F^2=25$ and $\lambda=50$, which we adopted for all the datasets with $K\geq 100$.
For datasets with fewer clusters (ImageNet 50, 
 Oxford-IIIT Pets, STL~10, CIFAR 10) we used a smaller embedding $d=32$. 
 
\section{Ablation Study on $\mathcal{L}_{cls}$ weight $\lambda$}
\begin{figure}[ht!]
    \centering

    \begin{minipage}[b]{\columnwidth} 
        \centering
        \includegraphics[width=0.8\columnwidth]{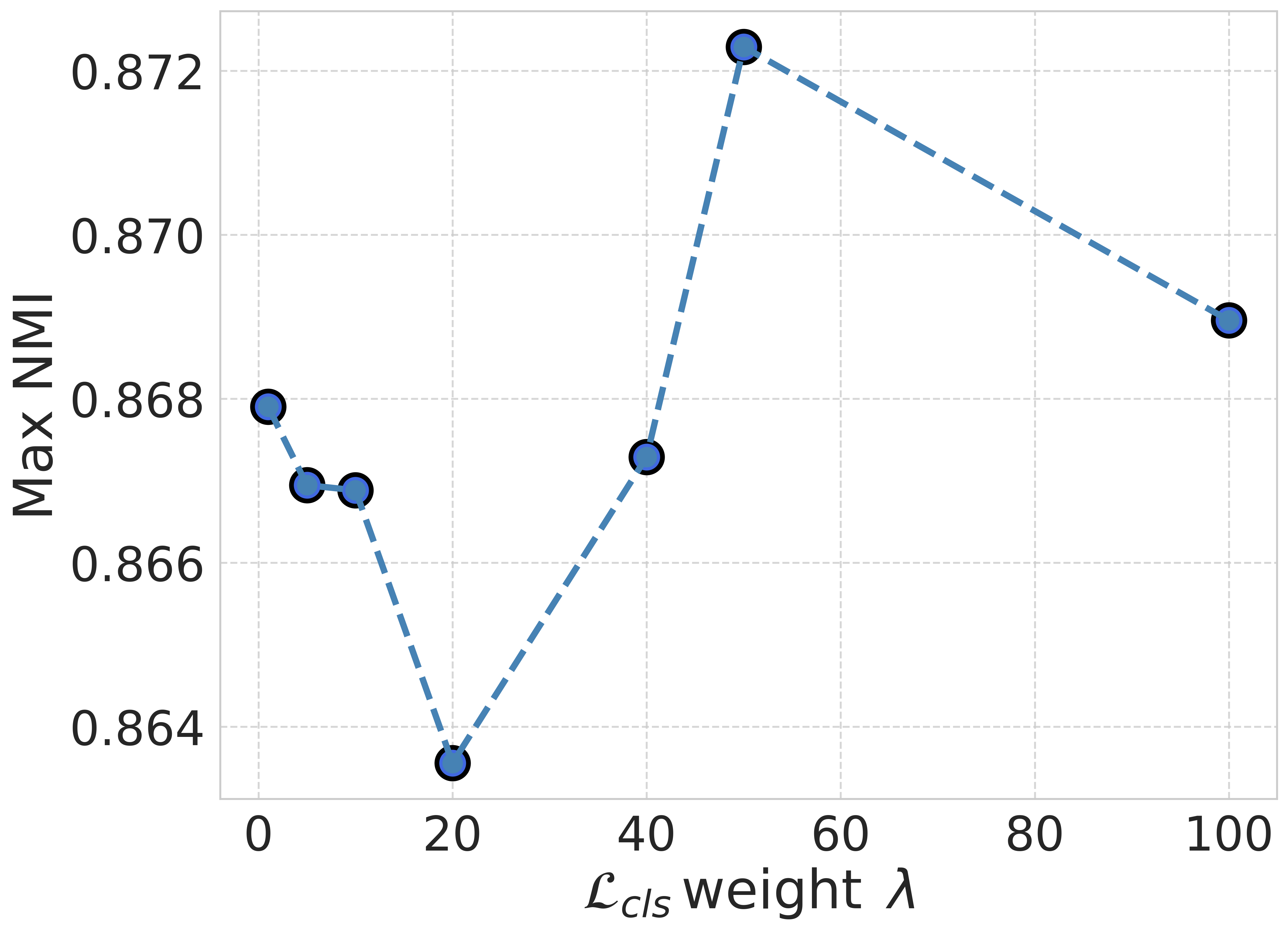}
        \label{fig:nmi_lambda}
    \end{minipage}
    \vspace{1em} 

    \begin{minipage}[b]{\columnwidth}
        \centering
        \includegraphics[width=0.8\columnwidth]{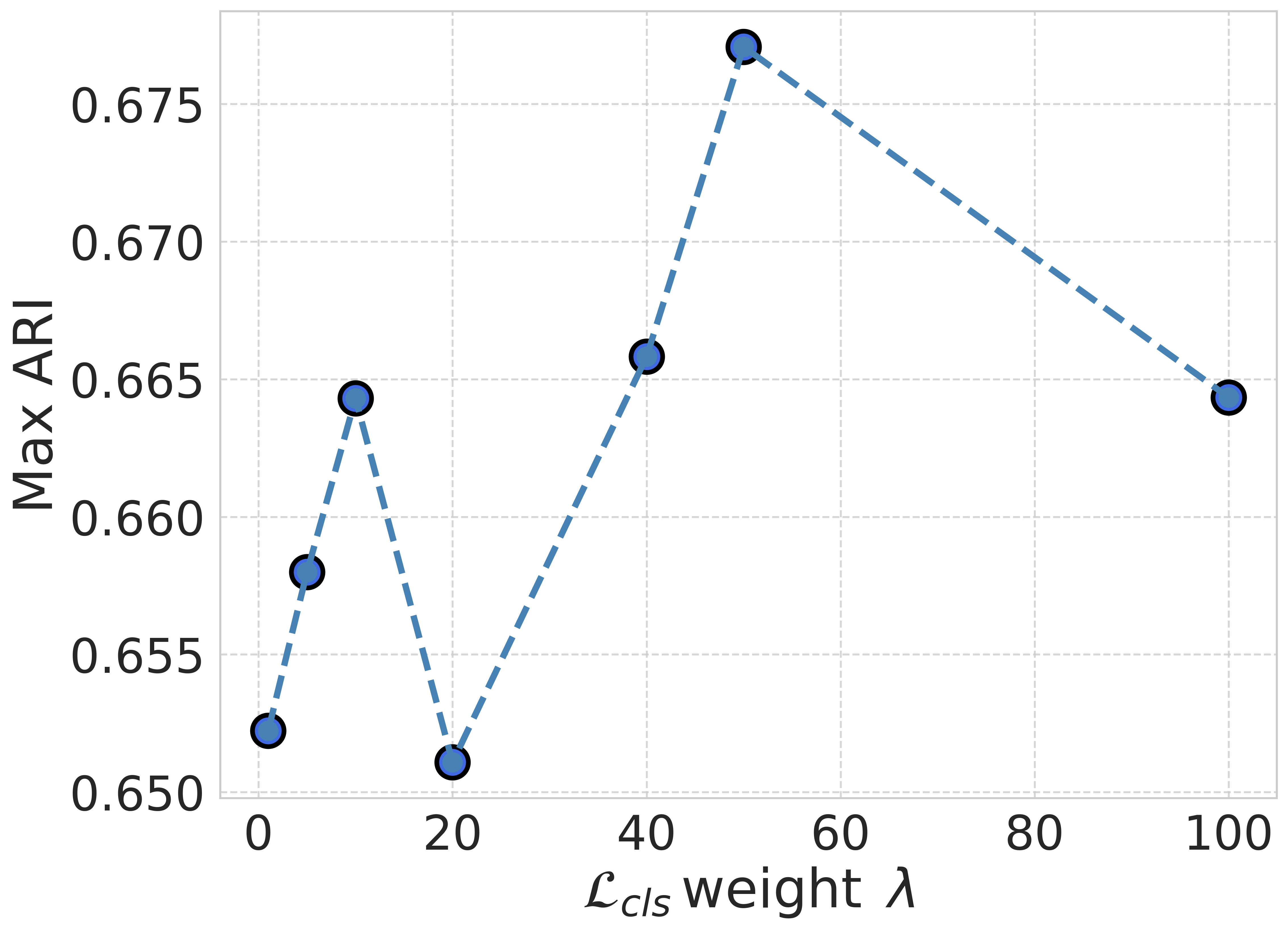}
        \label{fig:ari_lambda}
    \end{minipage}

    \vspace{1em}
    \caption{Clustering performance metrics (NMI, ACC, ARI) across $\lambda$ values for embedding dimension $d = 64$. Each image corresponds to a specific metric: (a) NMI, (b) ACC, and (c) ARI.}
    \label{fig:metrics_lambda}
\end{figure}

In \autoref{fig:metrics_lambda} 
we show the impact of varying the weight $\lambda$ associated with the classification loss $\mathcal{L}_{cls}$ on the overall clustering performance. 

\section{Ablation Study on Rescaling Factor $F^2$ and Embedding Dimension $d$}
\begin{figure*}[t!]
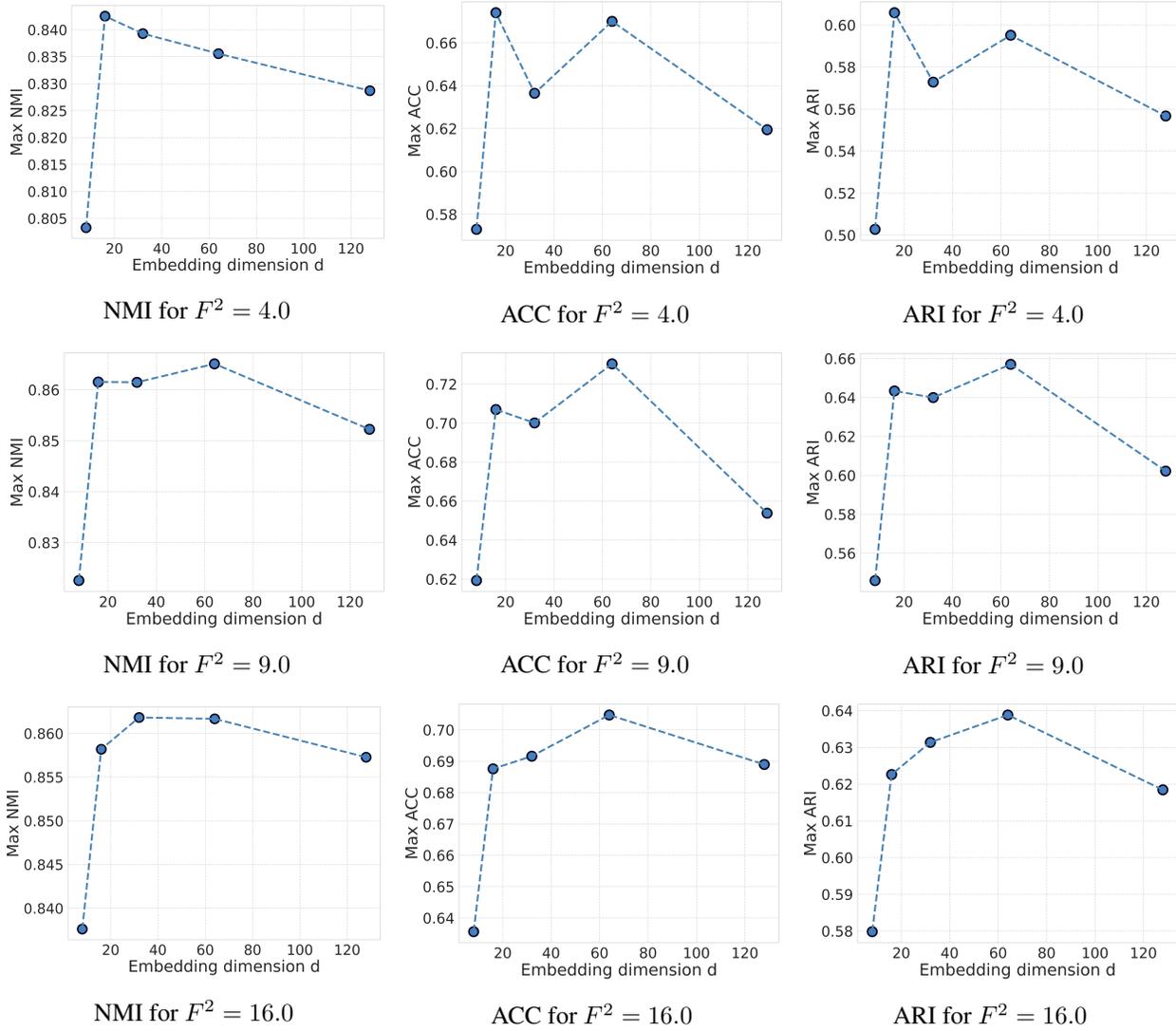

    \centering
    \displayFullPageImages{4.0}{9.0}{16.0}%
    {}
    \caption{Clustering performance metrics for rescaling factors 
    \( F^2 = 4.0, 9.0, 16.0 \) (top, middle, bottom, respectively) as a 
    function of the embedding dimension \( d \). Each column shows curves for a different clustering metric (NMI, ACC, and ARI).}
    \label{fig:rescaling_4_9_16}
\end{figure*}

In \autoref{fig:rescaling_4_9_16} and \autoref{fig:rescaling_25_36_49}
we explore the effect on clustering performance 
of varying the rescaling factor \( F^2 \) and 
the embedding dimension \( d \).

The noise rescaling factor \( F^2 \) plays a crucial role in our diffusion model by modulating the noise variance during the forward process. This parameter directly influences the model's ability to balance stability and exploration, which are essential for effective clustering. 

Insufficient noise (low \( F^2 \)) results in trivial solutions where embeddings fail to explore the latent space adequately. Conversely, excessive noise (high \( F^2 \)) destabilizes the learning process, disrupting inter-cluster separability. 

Our results, consistent with those in~\cite{gao2022difformer}  
for text generation, confirm the need to find an optimal value (for us \( F^2 = 25.0 \)) that strikes a balance between enabling the model to avoid degeneracy and preserving cluster coherence. In practical terms, our findings highlight that \( F^2 \) serves as a key hyperparameter for fine-tuning clustering models, with significant impacts on metrics such as NMI, ACC, and ARI.

\section{Clustering Visualization on ImageNet-50}
\begin{figure*}[t!]
    \centering
   \includegraphics[width=1.7\columnwidth]{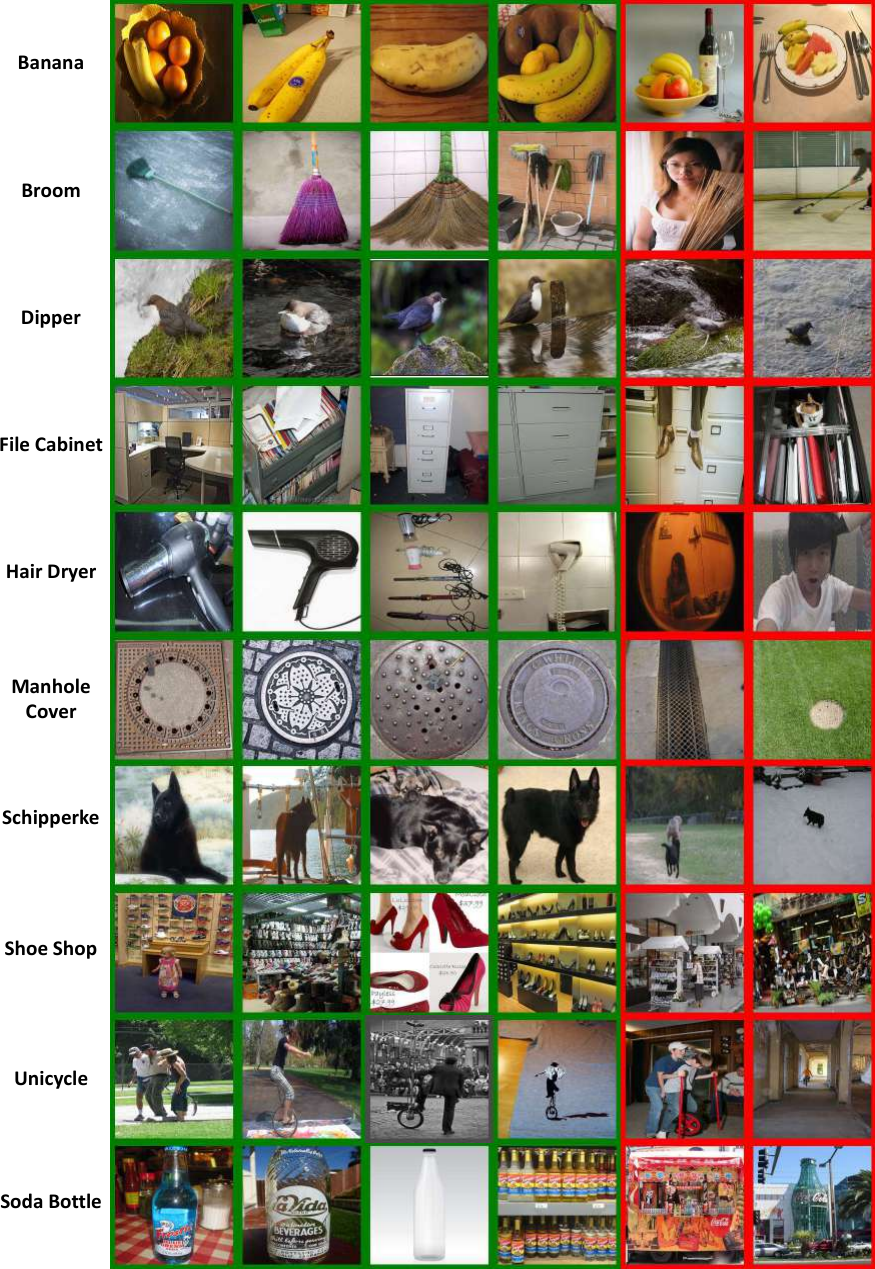}
  \caption{Visualization of supervised clustering results on the ImageNet 50 dataset. Each row corresponds to a single class, with the class name displayed on the left. For each class, the first four images represent the correctly classified samples with the highest confidence (outlined in green), while the last two images represent the incorrectly classified samples with the lowest confidence (outlined in red).}
    \label{fig:sup_clustering_1}
\end{figure*}

In \autoref{fig:sup_clustering_1} we provide a detailed visualization of the clustering results on the ImageNet-50 dataset. For ten different categories, we 
show images with highest and lowest confidence of belonging to each category. This illustrates the 
strength of our model and allows to visually explore the extent to which misclassifications are understandable. 

\begin{figure*}[t!]
    \centering
    \displayFullPageImages{25.0}{36.0}{49.0}%
    {}
    \caption{Clustering performance metrics for rescaling factors 
    \( F^2 = 25.0, 36.0, 49.0 \)
 (top, middle, bottom, respectively) as a 
    function of the embedding dimension \( d \). Each column shows curves for a different clustering metric (NMI, ACC, and ARI).}    
    \label{fig:rescaling_25_36_49}
\end{figure*}
\newpage

%% file: main.bbl
\begin{thebibliography}{68}
\providecommand{\natexlab}[1]{#1}
\providecommand{\url}[1]{\texttt{#1}}
\expandafter\ifx\csname urlstyle\endcsname\relax
  \providecommand{\doi}[1]{doi: #1}\else
  \providecommand{\doi}{doi: \begingroup \urlstyle{rm}\Url}\fi

\bibitem[Adaloglou et~al.(2023)Adaloglou, Michels, Kalisch, and Kollmann]{adaloglou2023exploring}
Adaloglou, N., Michels, F., Kalisch, H., and Kollmann, M.
\newblock Exploring the limits of deep image clustering using pretrained models.
\newblock In \emph{34th British Machine Vision Conference}, 2023.

\bibitem[Amrani et~al.(2022)Amrani, Karlinsky, and Bronstein]{amrani2022self}
Amrani, E., Karlinsky, L., and Bronstein, A.
\newblock Self-supervised classification network.
\newblock In \emph{European Conference on Computer Vision}, pp.\  116--132. Springer, 2022.

\bibitem[Balestriero et~al.(2023)Balestriero, Ibrahim, Sobal, Morcos, Shekhar, Goldstein, Bordes, Bardes, Mialon, Tian, Schwarzschild, Wilson, Geiping, Garrido, Fernandez, Bar, Pirsiavash, LeCun, and Goldblum]{balestriero2023cookbook}
Balestriero, R., Ibrahim, M., Sobal, V., Morcos, A., Shekhar, S., Goldstein, T., Bordes, F., Bardes, A., Mialon, G., Tian, Y., Schwarzschild, A., Wilson, A.~G., Geiping, J., Garrido, Q., Fernandez, P., Bar, A., Pirsiavash, H., LeCun, Y., and Goldblum, M.
\newblock A cookbook of self-supervised learning, 2023.

\bibitem[Ben-David(2018)]{ben2018clustering}
Ben-David, S.
\newblock Clustering-what both theoreticians and practitioners are doing wrong.
\newblock In \emph{Proceedings of the AAAI Conference on Artificial Intelligence}, volume~32, 2018.

\bibitem[Caron et~al.(2018)Caron, Bojanowski, Joulin, and Douze]{caron2018deep}
Caron, M., Bojanowski, P., Joulin, A., and Douze, M.
\newblock Deep clustering for unsupervised learning of visual features.
\newblock In \emph{Proceedings of the European conference on computer vision (ECCV)}, pp.\  132--149, 2018.

\bibitem[Caron et~al.(2021)Caron, Touvron, Misra, J{\'e}gou, Mairal, Bojanowski, and Joulin]{caron2021emerging}
Caron, M., Touvron, H., Misra, I., J{\'e}gou, H., Mairal, J., Bojanowski, P., and Joulin, A.
\newblock Emerging properties in self-supervised vision transformers.
\newblock In \emph{Proceedings of the IEEE/CVF international conference on computer vision}, pp.\  9650--9660, 2021.

\bibitem[Chan(2024)]{chan2024tutorial}
Chan, S.~H.
\newblock Tutorial on diffusion models for imaging and vision.
\newblock \emph{arXiv preprint arXiv:2403.18103}, 2024.

\bibitem[Chang \& Fisher~III(2013)Chang and Fisher~III]{chang2013parallel}
Chang, J. and Fisher~III, J.~W.
\newblock Parallel sampling of dp mixture models using sub-cluster splits.
\newblock \emph{Advances in Neural Information Processing Systems}, 26, 2013.

\bibitem[Chelly et~al.(2025)Chelly, Uziel, Freifeld, and Pakman]{chellyconsistent}
Chelly, I., Uziel, R., Freifeld, O., and Pakman, A.
\newblock Consistent amortized clustering via generative flow networks.
\newblock In \emph{The 28th International Conference on Artificial Intelligence and Statistics}, 2025.

\bibitem[Chen et~al.(2020)Chen, Kornblith, Norouzi, and Hinton]{chen2020simple}
Chen, T., Kornblith, S., Norouzi, M., and Hinton, G.
\newblock A simple framework for contrastive learning of visual representations.
\newblock In \emph{International conference on machine learning}, pp.\  1597--1607. PMLR, 2020.

\bibitem[Chen \& He(2021)Chen and He]{chen2021exploring}
Chen, X. and He, K.
\newblock Exploring simple siamese representation learning.
\newblock In \emph{Proceedings of the IEEE/CVF conference on computer vision and pattern recognition}, pp.\  15750--15758, 2021.

\bibitem[Chicco(2021)]{chicco2021siamese}
Chicco, D.
\newblock Siamese neural networks: An overview.
\newblock \emph{Artificial neural networks}, pp.\  73--94, 2021.

\bibitem[Coates et~al.(2011)Coates, Ng, and Lee]{coates2011analysis}
Coates, A., Ng, A., and Lee, H.
\newblock An analysis of single-layer networks in unsupervised feature learning.
\newblock In \emph{Proceedings of the fourteenth international conference on artificial intelligence and statistics}, pp.\  215--223. JMLR Workshop and Conference Proceedings, 2011.

\bibitem[Deng et~al.(2009)Deng, Dong, Socher, Li, Li, and Fei-Fei]{deng2009imagenet}
Deng, J., Dong, W., Socher, R., Li, L.-J., Li, K., and Fei-Fei, L.
\newblock Imagenet: A large-scale hierarchical image database.
\newblock In \emph{2009 IEEE conference on computer vision and pattern recognition}, pp.\  248--255. Ieee, 2009.

\bibitem[Dhariwal \& Nichol(2021)Dhariwal and Nichol]{dhariwal2021diffusion}
Dhariwal, P. and Nichol, A.
\newblock Diffusion models beat gans on image synthesis.
\newblock \emph{Advances in neural information processing systems}, 34:\penalty0 8780--8794, 2021.

\bibitem[Dieleman et~al.(2022)Dieleman, Sartran, Roshannai, Savinov, Ganin, Richemond, Doucet, Strudel, Dyer, Durkan, et~al.]{dieleman2022continuous}
Dieleman, S., Sartran, L., Roshannai, A., Savinov, N., Ganin, Y., Richemond, P.~H., Doucet, A., Strudel, R., Dyer, C., Durkan, C., et~al.
\newblock Continuous diffusion for categorical data.
\newblock \emph{arXiv preprint arXiv:2211.15089}, 2022.

\bibitem[Dinari et~al.(2019)Dinari, Yu, Freifeld, and Fisher]{dinari2019distributed}
Dinari, O., Yu, A., Freifeld, O., and Fisher, J.
\newblock Distributed mcmc inference in dirichlet process mixture models using julia.
\newblock In \emph{2019 19th IEEE/ACM International Symposium on Cluster, Cloud and Grid Computing (CCGRID)}, pp.\  518--525. IEEE, 2019.

\bibitem[Dosovitskiy(2020)]{dosovitskiy2020image}
Dosovitskiy, A.
\newblock An image is worth 16x16 words: Transformers for image recognition at scale.
\newblock \emph{arXiv preprint arXiv:2010.11929}, 2020.

\bibitem[Ericsson et~al.(2021)Ericsson, Gouk, and Hospedales]{ericsson2021well}
Ericsson, L., Gouk, H., and Hospedales, T.~M.
\newblock How well do self-supervised models transfer?
\newblock In \emph{Proceedings of the IEEE/CVF conference on computer vision and pattern recognition}, pp.\  5414--5423, 2021.

\bibitem[Fahad et~al.(2014)Fahad, Alshatri, Tari, Alamri, Khalil, Zomaya, Foufou, and Bouras]{fahad:2014:NMI_ARI}
Fahad, A., Alshatri, N., Tari, Z., Alamri, A., Khalil, I., Zomaya, A.~Y., Foufou, S., and Bouras, A.
\newblock A survey of clustering algorithms for big data: Taxonomy and empirical analysis.
\newblock \emph{IEEE transactions on emerging topics in computing}, 2\penalty0 (3):\penalty0 267--279, 2014.

\bibitem[Fei-Fei et~al.(2004)Fei-Fei, Fergus, and Perona]{fei2004learning}
Fei-Fei, L., Fergus, R., and Perona, P.
\newblock Learning generative visual models from few training examples: An incremental bayesian approach tested on 101 object categories.
\newblock In \emph{2004 conference on computer vision and pattern recognition workshop}, pp.\  178--178. IEEE, 2004.

\bibitem[Friebel et~al.(2022)Friebel, Johann, Drasdo, and Hoehme]{friebel2022guided}
Friebel, A., Johann, T., Drasdo, D., and Hoehme, S.
\newblock Guided interactive image segmentation using machine learning and color-based image set clustering.
\newblock \emph{Bioinformatics}, 38\penalty0 (19):\penalty0 4622--4628, 2022.

\bibitem[Gao et~al.(2024)Gao, Guo, Tan, Zhu, Zhang, Bian, and Xu]{gao2022difformer}
Gao, Z., Guo, J., Tan, X., Zhu, Y., Zhang, F., Bian, J., and Xu, L.
\newblock Empowering diffusion models on the embedding space for text generation.
\newblock In \emph{Proceedings of the 2024 Conference of the North American Chapter of the Association for Computational Linguistics: Human Language Technologies (Volume 1: Long Papers)}, pp.\  4664--4683, 2024.

\bibitem[Gong et~al.(2022)Gong, Li, Feng, Wu, and Kong]{gong2022diffuseq}
Gong, S., Li, M., Feng, J., Wu, Z., and Kong, L.
\newblock Diffuseq: Sequence to sequence text generation with diffusion models.
\newblock \emph{arXiv preprint arXiv:2210.08933}, 2022.

\bibitem[Grill et~al.(2020)Grill, Strub, Altch{\'e}, Tallec, Richemond, Buchatskaya, Doersch, Avila~Pires, Guo, Gheshlaghi~Azar, et~al.]{grill2020bootstrap}
Grill, J.-B., Strub, F., Altch{\'e}, F., Tallec, C., Richemond, P., Buchatskaya, E., Doersch, C., Avila~Pires, B., Guo, Z., Gheshlaghi~Azar, M., et~al.
\newblock Bootstrap your own latent-a new approach to self-supervised learning.
\newblock \emph{Advances in neural information processing systems}, 33:\penalty0 21271--21284, 2020.

\bibitem[Gui et~al.(2024)Gui, Chen, Zhang, Cao, Sun, Luo, and Tao]{gui2024survey}
Gui, J., Chen, T., Zhang, J., Cao, Q., Sun, Z., Luo, H., and Tao, D.
\newblock A survey on self-supervised learning: Algorithms, applications, and future trends.
\newblock \emph{IEEE Transactions on Pattern Analysis and Machine Intelligence}, 2024.

\bibitem[Halvagal et~al.(2023)Halvagal, Laborieux, and Zenke]{halvagal2023implicit}
Halvagal, M.~S., Laborieux, A., and Zenke, F.
\newblock Implicit variance regularization in non-contrastive ssl.
\newblock In \emph{Proceedings of the 37th International Conference on Neural Information Processing Systems}, pp.\  63409--63436, 2023.

\bibitem[Han et~al.(2020)Han, Park, Park, Kim, and Cha]{han2020mitigating}
Han, S., Park, S., Park, S., Kim, S., and Cha, M.
\newblock Mitigating embedding and class assignment mismatch in unsupervised image classification.
\newblock In \emph{European Conference on Computer Vision}, pp.\  768--784. Springer, 2020.

\bibitem[Hang et~al.(2023)Hang, Gu, Li, Bao, Chen, Hu, Geng, and Guo]{Hang_2023_ICCV}
Hang, T., Gu, S., Li, C., Bao, J., Chen, D., Hu, H., Geng, X., and Guo, B.
\newblock Efficient diffusion training via min-snr weighting strategy.
\newblock In \emph{Proceedings of the IEEE/CVF International Conference on Computer Vision (ICCV)}, pp.\  7441--7451, October 2023.

\bibitem[Ho et~al.(2020)Ho, Jain, and Abbeel]{ho2020denoising}
Ho, J., Jain, A., and Abbeel, P.
\newblock Denoising diffusion probabilistic models.
\newblock \emph{Advances in neural information processing systems}, 33:\penalty0 6840--6851, 2020.

\bibitem[Hospedales et~al.(2021)Hospedales, Antoniou, Micaelli, and Storkey]{hospedales2021meta}
Hospedales, T., Antoniou, A., Micaelli, P., and Storkey, A.
\newblock Meta-learning in neural networks: A survey.
\newblock \emph{IEEE transactions on pattern analysis and machine intelligence}, 44\penalty0 (9):\penalty0 5149--5169, 2021.

\bibitem[Huang et~al.(2022)Huang, Chen, Zhang, and Shan]{huang2022learning}
Huang, Z., Chen, J., Zhang, J., and Shan, H.
\newblock Learning representation for clustering via prototype scattering and positive sampling.
\newblock \emph{IEEE Transactions on Pattern Analysis and Machine Intelligence}, 2022.

\bibitem[Jaiswal et~al.(2020)Jaiswal, Babu, Zadeh, Banerjee, and Makedon]{jaiswal2020survey}
Jaiswal, A., Babu, A.~R., Zadeh, M.~Z., Banerjee, D., and Makedon, F.
\newblock A survey on contrastive self-supervised learning.
\newblock \emph{Technologies}, 9\penalty0 (1):\penalty0 2, 2020.

\bibitem[Ji et~al.(2019)Ji, Henriques, and Vedaldi]{ji2019invariant}
Ji, X., Henriques, J.~F., and Vedaldi, A.
\newblock Invariant information clustering for unsupervised image classification and segmentation.
\newblock In \emph{Proceedings of the IEEE/CVF international conference on computer vision}, pp.\  9865--9874, 2019.

\bibitem[Jiang et~al.(2016)Jiang, Zheng, Tan, Tang, and Zhou]{jiang2016variational}
Jiang, Z., Zheng, Y., Tan, H., Tang, B., and Zhou, H.
\newblock Variational deep embedding: An unsupervised and generative approach to clustering.
\newblock \emph{arXiv preprint arXiv:1611.05148}, 2016.

\bibitem[Jurewicz et~al.(2023)Jurewicz, Taylor, and Derczynski]{jurewicz2023catalog}
Jurewicz, M.~M., Taylor, G.~W., and Derczynski, L.
\newblock The catalog problem: clustering and ordering variable-sized sets.
\newblock In \emph{International Conference on Machine Learning}, pp.\  15528--15545. PMLR, 2023.

\bibitem[Karim et~al.(2021)Karim, Beyan, Zappa, Costa, Rebholz-Schuhmann, Cochez, and Decker]{karim2021deep}
Karim, M.~R., Beyan, O., Zappa, A., Costa, I.~G., Rebholz-Schuhmann, D., Cochez, M., and Decker, S.
\newblock Deep learning-based clustering approaches for bioinformatics.
\newblock \emph{Briefings in bioinformatics}, 22\penalty0 (1):\penalty0 393--415, 2021.

\bibitem[Krizhevsky et~al.(2009)Krizhevsky, Hinton, et~al.]{krizhevsky2009learning}
Krizhevsky, A., Hinton, G., et~al.
\newblock Learning multiple layers of features from tiny images.
\newblock 2009.

\bibitem[Li et~al.(2022)Li, Thickstun, Gulrajani, Liang, and Hashimoto]{li2022diffusion}
Li, X., Thickstun, J., Gulrajani, I., Liang, P.~S., and Hashimoto, T.~B.
\newblock Diffusion-lm improves controllable text generation.
\newblock \emph{Advances in Neural Information Processing Systems}, 35:\penalty0 4328--4343, 2022.

\bibitem[Liu et~al.(2022)Liu, Suganuma, and Okatani]{liu2022bridging}
Liu, K.-J., Suganuma, M., and Okatani, T.
\newblock Bridging the gap from asymmetry tricks to decorrelation principles in non-contrastive self-supervised learning.
\newblock \emph{Advances in Neural Information Processing Systems}, 35:\penalty0 19824--19835, 2022.

\bibitem[Luo(2022)]{luo2022understanding}
Luo, C.
\newblock Understanding diffusion models: A unified perspective.
\newblock \emph{arXiv preprint arXiv:2208.11970}, 2022.

\bibitem[Mittal et~al.(2022)Mittal, Pandey, Saraswat, Kumar, Pal, and Modwel]{mittal2022comprehensive}
Mittal, H., Pandey, A.~C., Saraswat, M., Kumar, S., Pal, R., and Modwel, G.
\newblock A comprehensive survey of image segmentation: clustering methods, performance parameters, and benchmark datasets.
\newblock \emph{Multimedia Tools and Applications}, pp.\  1--26, 2022.

\bibitem[Nakkiran et~al.(2024)Nakkiran, Bradley, Zhou, and Advani]{nakkiran2024step}
Nakkiran, P., Bradley, A., Zhou, H., and Advani, M.
\newblock Step-by-step diffusion: An elementary tutorial.
\newblock \emph{arXiv preprint arXiv:2406.08929}, 2024.

\bibitem[Nichol \& Dhariwal(2021)Nichol and Dhariwal]{nichol2021improved}
Nichol, A.~Q. and Dhariwal, P.
\newblock Improved denoising diffusion probabilistic models.
\newblock In \emph{International conference on machine learning}, pp.\  8162--8171. PMLR, 2021.

\bibitem[Nilsback \& Zisserman(2008)Nilsback and Zisserman]{nilsback2008automated}
Nilsback, M.-E. and Zisserman, A.
\newblock Automated flower classification over a large number of classes.
\newblock In \emph{2008 Sixth Indian conference on computer vision, graphics \& image processing}, pp.\  722--729. IEEE, 2008.

\bibitem[{\"O}zbulak et~al.(2023){\"O}zbulak, Lee, Boga, Anzaku, Park, Van~Messem, De~Neve, and Vankerschaver]{ozbulak2023know}
{\"O}zbulak, U., Lee, H.~J., Boga, B., Anzaku, E.~T., Park, H., Van~Messem, A., De~Neve, W., and Vankerschaver, J.
\newblock Know your self-supervised learning: a survey on image-based generative and discriminative training.
\newblock \emph{TRANSACTIONS ON MACHINE LEARNING RESEARCH}, 2023.

\bibitem[Pakman et~al.(2020)Pakman, Wang, Mitelut, Lee, and Paninski]{pakman2020neural}
Pakman, A., Wang, Y., Mitelut, C., Lee, J., and Paninski, L.
\newblock Neural clustering processes.
\newblock In \emph{International Conference on Machine Learning}, pp.\  7455--7465. PMLR, 2020.

\bibitem[Parkhi et~al.(2012)Parkhi, Vedaldi, Zisserman, and Jawahar]{parkhi2012cats}
Parkhi, O.~M., Vedaldi, A., Zisserman, A., and Jawahar, C.
\newblock Cats and dogs.
\newblock In \emph{2012 IEEE conference on computer vision and pattern recognition}, pp.\  3498--3505. IEEE, 2012.

\bibitem[Ren et~al.(2024)Ren, Pu, Yang, Xu, Li, Pu, Philip, and He]{ren2024deep}
Ren, Y., Pu, J., Yang, Z., Xu, J., Li, G., Pu, X., Philip, S.~Y., and He, L.
\newblock Deep clustering: A comprehensive survey.
\newblock \emph{IEEE Transactions on Neural Networks and Learning Systems}, 2024.

\bibitem[Richemond et~al.(2023)Richemond, Tam, Tang, Strub, Piot, and Hill]{richemond2023edge}
Richemond, P.~H., Tam, A., Tang, Y., Strub, F., Piot, B., and Hill, F.
\newblock The edge of orthogonality: A simple view of what makes byol tick.
\newblock In \emph{International Conference on Machine Learning}, pp.\  29063--29081. PMLR, 2023.

\bibitem[Rombach et~al.(2022)Rombach, Blattmann, Lorenz, Esser, and Ommer]{rombach2022high}
Rombach, R., Blattmann, A., Lorenz, D., Esser, P., and Ommer, B.
\newblock High-resolution image synthesis with latent diffusion models.
\newblock In \emph{Proceedings of the IEEE/CVF conference on computer vision and pattern recognition}, pp.\  10684--10695, 2022.

\bibitem[Ronen et~al.(2022)Ronen, Finder, and Freifeld]{Ronen:CVPR:2022:DeepDPM}
Ronen, M., Finder, S.~E., and Freifeld, O.
\newblock Deep{DPM}: Deep clustering with an unknown number of clusters.
\newblock In \emph{Proceedings of the IEEE/CVF Conference on Computer Vision and Pattern Recognition}, pp.\  9861--9870, 2022.

\bibitem[Salimans \& Ho(2022)Salimans and Ho]{salimansprogressive}
Salimans, T. and Ho, J.
\newblock Progressive distillation for fast sampling of diffusion models.
\newblock In \emph{International Conference on Learning Representations}, 2022.

\bibitem[Shwartz~Ziv \& LeCun(2024)Shwartz~Ziv and LeCun]{shwartz2024compress}
Shwartz~Ziv, R. and LeCun, Y.
\newblock To compress or not to compress—self-supervised learning and information theory: A review.
\newblock \emph{Entropy}, 26\penalty0 (3):\penalty0 252, 2024.

\bibitem[Sohl-Dickstein et~al.(2015)Sohl-Dickstein, Weiss, Maheswaranathan, and Ganguli]{sohl2015deep}
Sohl-Dickstein, J., Weiss, E., Maheswaranathan, N., and Ganguli, S.
\newblock Deep unsupervised learning using nonequilibrium thermodynamics.
\newblock In \emph{International conference on machine learning}, pp.\  2256--2265. PMLR, 2015.

\bibitem[Song et~al.(2021{\natexlab{a}})Song, Li, and Liu]{song2021deep}
Song, H., Li, P., and Liu, H.
\newblock Deep clustering based fair outlier detection.
\newblock In \emph{Proceedings of the 27th ACM SIGKDD Conference on Knowledge Discovery \& Data Mining}, pp.\  1481--1489, 2021{\natexlab{a}}.

\bibitem[Song et~al.(2021{\natexlab{b}})Song, Meng, and Ermon]{song2020denoising}
Song, J., Meng, C., and Ermon, S.
\newblock Denoising diffusion implicit models.
\newblock In \emph{International Conference on Learning Representations}, 2021{\natexlab{b}}.

\bibitem[Song et~al.(2021{\natexlab{c}})Song, Sohl-Dickstein, Kingma, Kumar, Ermon, and Poole]{songscore}
Song, Y., Sohl-Dickstein, J., Kingma, D.~P., Kumar, A., Ermon, S., and Poole, B.
\newblock Score-based generative modeling through stochastic differential equations.
\newblock In \emph{International Conference on Learning Representations}, 2021{\natexlab{c}}.

\bibitem[Tao et~al.(2022)Tao, Wang, Zhu, Dong, Song, Huang, and Dai]{tao2022exploring}
Tao, C., Wang, H., Zhu, X., Dong, J., Song, S., Huang, G., and Dai, J.
\newblock Exploring the equivalence of siamese self-supervised learning via a unified gradient framework.
\newblock In \emph{Proceedings of the IEEE/CVF Conference on Computer Vision and Pattern Recognition}, pp.\  14431--14440, 2022.

\bibitem[Tian et~al.(2021)Tian, Chen, and Ganguli]{tian2021understanding}
Tian, Y., Chen, X., and Ganguli, S.
\newblock Understanding self-supervised learning dynamics without contrastive pairs.
\newblock In \emph{International Conference on Machine Learning}, pp.\  10268--10278. PMLR, 2021.

\bibitem[Turner et~al.(2024)Turner, Diaconu, Markou, Shysheya, Foong, and Mlodozeniec]{turner2024denoising}
Turner, R.~E., Diaconu, C.-D., Markou, S., Shysheya, A., Foong, A.~Y., and Mlodozeniec, B.
\newblock Denoising diffusion probabilistic models in six simple steps.
\newblock \emph{arXiv preprint arXiv:2402.04384}, 2024.

\bibitem[Van~Gansbeke et~al.(2020)Van~Gansbeke, Vandenhende, Georgoulis, Proesmans, and Van~Gool]{van2020scan}
Van~Gansbeke, W., Vandenhende, S., Georgoulis, S., Proesmans, M., and Van~Gool, L.
\newblock {SCAN: Learning to classify images without labels}.
\newblock In \emph{European conference on computer vision}, pp.\  268--285. Springer, 2020.

\bibitem[Wang et~al.(2021)Wang, Chen, Du, and Tian]{wang2021towards}
Wang, X., Chen, X., Du, S.~S., and Tian, Y.
\newblock Towards demystifying representation learning with non-contrastive self-supervision.
\newblock \emph{arXiv preprint arXiv:2110.04947}, 2021.

\bibitem[Wang et~al.(2024)Wang, Lee, Basu, Lee, Teh, Paninski, and Pakman]{wang2024amortized}
Wang, Y., Lee, Y., Basu, P., Lee, J., Teh, Y.~W., Paninski, L., and Pakman, A.
\newblock Amortized probabilistic detection of communities in graphs.
\newblock \emph{Structured Probabilistic Inference \& Generative Modeling workshop at ICML}, 2024.

\bibitem[Wei et~al.(2024)Wei, Zhang, Huang, and Zhou]{wei2024overview}
Wei, X., Zhang, Z., Huang, H., and Zhou, Y.
\newblock An overview on deep clustering.
\newblock \emph{Neurocomputing}, pp.\  127761, 2024.

\bibitem[Xie et~al.(2016)Xie, Girshick, and Farhadi]{xie2016unsupervised}
Xie, J., Girshick, R., and Farhadi, A.
\newblock Unsupervised deep embedding for clustering analysis.
\newblock In \emph{International conference on machine learning}, pp.\  478--487. PMLR, 2016.

\bibitem[Zhou et~al.(2024)Zhou, Xu, Zheng, Chen, Li, Bu, Wu, Wang, Zhu, and Ester]{zhou:2022:survey1}
Zhou, S., Xu, H., Zheng, Z., Chen, J., Li, Z., Bu, J., Wu, J., Wang, X., Zhu, W., and Ester, M.
\newblock A comprehensive survey on deep clustering: Taxonomy, challenges, and future directions.
\newblock \emph{ACM Computing Surveys}, 2024.

\bibitem[Zhou \& Zhang(2022)Zhou and Zhang]{zhou2022deep}
Zhou, X. and Zhang, N.~L.
\newblock Deep clustering with features from self-supervised pretraining.
\newblock \emph{arXiv preprint arXiv:2207.13364}, 2022.

\end{thebibliography}
